\documentclass{ecai} 

\usepackage{latexsym}
\usepackage{amssymb}
\usepackage{amsmath}
\usepackage{amsthm}
\usepackage{booktabs}
\usepackage{enumitem}
\usepackage{graphicx}
\usepackage{color}

\newcommand{\BibTeX}{B\kern-.05em{\sc i\kern-.025em b}\kern-.08em\TeX}

\usepackage[hidelinks]{hyperref}
\usepackage{color}

\usepackage[capitalize]{cleveref}
\crefname{section}{Sec.}{Secs.}
\Crefname{section}{Section}{Sections}
\Crefname{table}{Table}{Tables}
\Crefname{algocf}{Algorithm}{Algorithms}
\crefname{table}{Tab.}{Tabs.}
\crefname{algocf}{Alg.}{Algs.}
\usepackage{multirow}
\usepackage{array}
\usepackage[ruled,vlined]{algorithm2e}
\usepackage{wrapfig} 
\usepackage{caption}
\usepackage[font=tiny]{subcaption}
\usepackage[export]{adjustbox}

\graphicspath{ {./figures/} }

\begin{document}


\begin{frontmatter}

\paperid{8969}

\title{IGAff: Benchmarking Adversarial Iterative and Genetic Affine Algorithms on Deep Neural Networks}

\author[A]{\fnms{Sebastian-Vasile}~\snm{Echim}}
\author[A]{\fnms{Andrei-Alexandru}~\snm{Preda}} 
\author[A]{\fnms{Dumitru-Clementin}~\snm{Cercel}\thanks{Corresponding author: dumitru.cercel@upb.ro.}}
\author[A,B]{\fnms{Florin}~\snm{Pop}}

\address[A]{Faculty of Automatic Control and Computers, National University of Science and Technology POLITEHNICA Bucharest, Bucharest, Romania}
\address[B]{National Institute for Research \& Development in Informatics - ICI Bucharest, Bucharest, Romania}

\begin{abstract}
Deep neural networks currently dominate many fields of the artificial intelligence landscape, achieving state-of-the-art results on numerous tasks while remaining hard to understand and exhibiting surprising weaknesses. An active area of research focuses on adversarial attacks, which aim to generate inputs that uncover these weaknesses. However, this proves challenging, especially in the black-box scenario where model details are inaccessible. This paper explores in detail the impact of such adversarial algorithms on ResNet-18, DenseNet-121, Swin Transformer V2, and Vision Transformer network architectures. Leveraging the Tiny ImageNet, Caltech-256, and Food-101 datasets, we benchmark two novel black-box iterative adversarial algorithms based on affine transformations and genetic algorithms: 1) Affine Transformation Attack (ATA), an iterative algorithm maximizing our attack score function using random affine transformations, and 2) Affine Genetic Attack (AGA), a genetic algorithm that involves random noise and affine transformations. We evaluate the performance of the models in the algorithm parameter variation, data augmentation, and global and targeted attack configurations. We also compare our algorithms with two black-box adversarial algorithms, Pixle and Square Attack. Our experiments yield better results on the image classification task than similar methods in the literature, achieving an accuracy improvement of up to 8.82\%. We provide noteworthy insights into successful adversarial defenses and attacks at both global and targeted levels, and demonstrate adversarial robustness through algorithm parameter variation.
\end{abstract}

\end{frontmatter}


\section{Introduction}

The robustness of deep learning models has been a topic of interest for decades \cite{seltzer2013robust,li2015robustpatients,you2015robustsentimentimage}, particularly since many are considered challenging to explain. Adversarial attacks are an equally relevant research area \cite{szegedy2013intriguing,gu2015robustadversarial,goodfellow2015explaining}, with various attack methods that efficiently find weaknesses even in the most promising models such as the transformer architecture, which has achieved state-of-the-art performance in recent years, including the field of computer vision through the Vision Transformer (ViT) \cite{dosovitskiy2021vit} and its derivatives. Considering the robustness challenges currently present in the novel deep neural network architectures, it is worth investigating the robustness of these new models against adversarial attacks.

Adversarial samples refer to crafted inputs that confuse deep learning models without a significant deviation from the data distribution \cite{szegedy2013intriguing}. The examples commonly described originate in the computer vision domain, where it has been proven that applying humanly imperceptible noise to images causes machine learning models to misclassify them~\cite{carlini2017towards}. Deep learning models are often perceived as black boxes, making them difficult to interpret and understand; thus, attacks not only have the opportunity to uncover weaknesses exhibited by models, but can also help train more robust model versions. 

Genetic algorithms (GAs), also known as evolutionary algorithms, were first adopted in the early 1970s~\cite{holland1992adaptation} and are part of an older trend of computational simulation of natural processes. 
Their search solution framework is largely problem-agnostic, allowing GAs to solve a wide range of problems, including more modern approaches~\cite{10.1162/evco_a_00282,stanley2002efficient}. GAs have been adapted in the past for many tasks, including training neural networks~\cite{szegedi2019evolutionary}, searching for neural architectures~\cite{10.1162/evco_a_00282}, or optimizing hyperparameters~\cite{996017}.
More recently, they have proved to be an efficient way of generating adversarial attacks~\cite{alzantot-etal-2018-generating,alzantot2019genattack,carlini2017towards,DBLP:journals/corr/SzegedyZSBEGF13} by combining and mutating candidate inputs that appear adversarially promising, demonstrating effectiveness without access to the model architecture or weights, which are often inaccessible. The use of adversarial genetic algorithms has also expanded to the natural language domain~\cite{alzantot2019genattack}, cybersecurity~\cite{rathore2023evoaattack}, the tabular setting~\cite{10191808}, and even being combined with reinforcement learning~\cite{10097974}.

\begin{figure}[!tb]
  \centering
  \includegraphics[width=1.005\columnwidth]{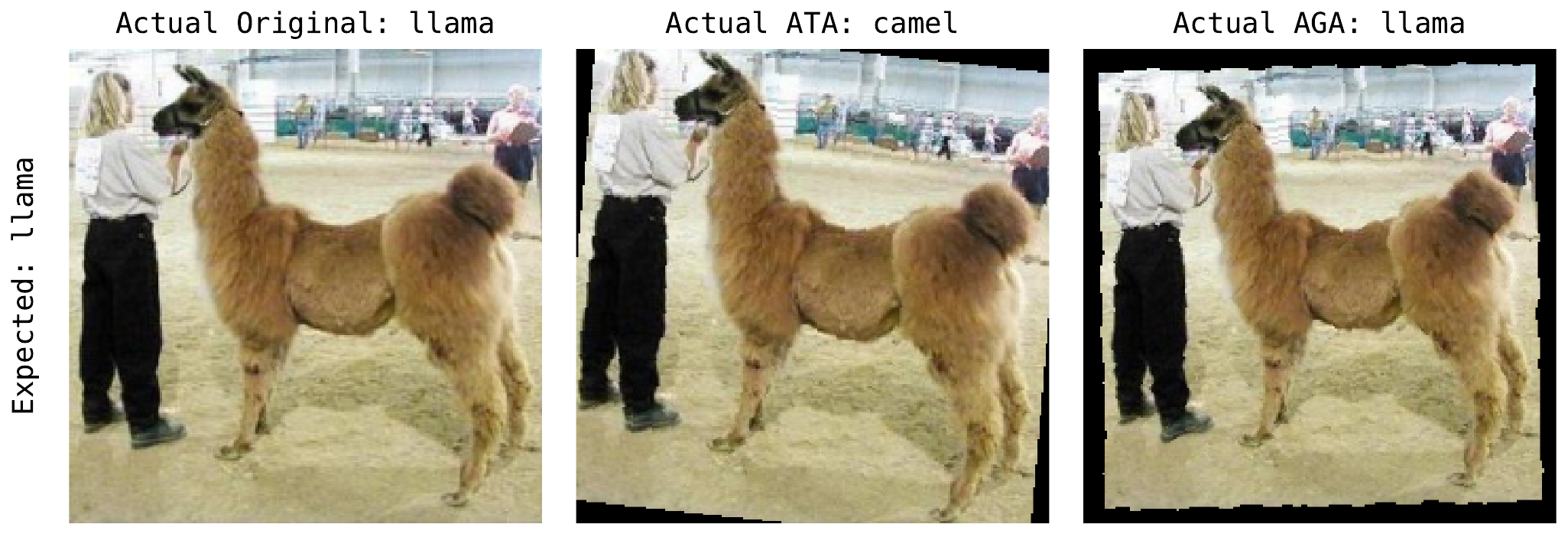}
  \caption{Affine Transformation Attack (ATA) and Affine Genetic Attack (AGA) output compared to the original image. } 
   \label{fig:ata-aga-examples}
\end{figure}

We assess the robustness of deep learning models by introducing two novel black-box iterative adversarial algorithms based on affine transformations and genetic algorithms: Affine Transformation Attack (ATA), an iterative adversarial method that applies affine transformations, and Affine Genetic Attack (AGA), a genetic algorithm that involves affine transformations and random noise. We evaluate our algorithms through comprehensive benchmark experiments for adversarial attack, defense, and data augmentation setups on Caltech-256 \cite{griffin2022caltech}, Food-101 \cite{bossard2014food}, and Tiny ImageNet (Tiny-ImageNet-200) \cite{russakovsky2015imagenet} datasets, two Convolutional Neural Network (CNN) architectures (ResNet-18 \cite{he2016resnet}, DenseNet-121 \cite{huang2017densenet}) and two computer vision transformer architectures (Swin Transformer V2 \cite{liu2022swinV2}, ViT \cite{dosovitskiy2021vit}).

The contributions of this paper are outlined as follows:
\begin{itemize}
   \item We introduce and thoroughly benchmark two novel black-box iterative adversarial algorithms based on affine transformations and genetic algorithms.
   \item We evaluate the data augmentation performance of the algorithms, gaining added value for more training data.
   \item We assess adversarial untargeted, global attacks and achieve outstanding results for defended and undefended attacks.
   \item We study adversarial targeted attacks on ten classes for a dataset and obtain improved results in targeted classification, enabling training possibilities for pre-trained models with particular class confusion.
   \item We vary the parameters of our algorithms and demonstrate the effectiveness of attacks with different parameter values.
   \item We qualitatively compare our algorithms with the Pixle and Square Attack black-box approaches.
\end{itemize}


\section{Related work}

\subsection{Adversarial Robustness}

Robustness is a quality that indicates the capacity of a model to generalize well to new or out-of-distribution samples. 
In the past, numerous methods have been used to obtain inputs that stretch models' abilities, from applying simple affine transformations on regular images~\cite{marchisio2023robcaps} to using datasets that feature spurious correlations~\cite{10.1007/s11263-023-01916-5}, and even generating adversarial inputs for specific models through attacks~\cite{marchisio2023robcaps,DBLP:journals/tmlr/ShaoSYCH22}. This approach refers to a wide range of methods, from white-box attacks, that take advantage of knowledge of the model parameters, such as the Fast Gradient Sign Method (FGSM)~\cite{goodfellow2015explaining} and Projected Gradient Descent (PGD)~\cite{madry2017towards}, to black-box attacks, like Carlini \& Wagner (C\&W)~\cite{carlini2017towards} and Pixle~\cite{Pomponi_2022}.

Based on adversarial experiments present in the literature, the robustness of computer vision models has been a topic of research for several years, among which the Convolutional Neural Networks have been studied in detail~\cite{10.1007/s11263-023-01916-5}, while the robustness of the Vision Transformer~\cite{dosovitskiy2021vit} has only been investigated more recently. 
The application of the transformer architecture to vision tasks has yielded state-of-the-art results in numerous studies, demonstrating improved robustness in specific settings.
Compared to convolutional models, ViTs rely less on high-frequency features~\cite{DBLP:journals/tmlr/ShaoSYCH22}, and, when pre-trained on large enough datasets, they generalize better against spurious correlations present in the data~\cite{10.1007/s11263-023-01916-5}. Shao et al. \cite{DBLP:journals/tmlr/ShaoSYCH22} show that this robustness is not tied to the attention mechanism because even CNN architectures become more robust when borrowing techniques from transformers, such as larger kernel sizes or invertible bottlenecks. One downside of Vision Transformers appears to be the lower transferability of attacks~\cite{DBLP:journals/tmlr/ShaoSYCH22}. However, recent work has improved their adversarial transferability limitation by leveraging the block-based architecture of Naseer et al.~\cite {DBLP:conf/iclr/NaseerR0KP22}, which employs the output of multiple intermediate transformer blocks instead of just the final one.

\subsection{Adversarial Affine Transformations}

Affine transformations refer to geometric transformations that preserve the parallelism of lines. Translations, rotations, scalings, and shears are often used to create synthetic augmentation data for many computer vision tasks \cite{athalye2018affine,shen2019affine}. This usage addresses situations where training data is scarce or diverse, affecting the most popular deep neural network architectures. Vision Transformers have been shown to require large amounts of data for effective pre-training, and a careful augmentation process can bring equal performance improvements in specific scenarios, given that a dataset is ten times larger in size~\cite{steiner2022how}. 

In recent work, Tian et al. \cite{tian2024affineattack} benchmarked simple image mutations such as affine transformations and bilateral, median, and Gaussian blur for model defense against FGSM, RFGSM (Randomized Fast Gradient Sign Method), and PGD attacks without increasing the initial training input size. Their results for the ImageNet dataset show that, in most cases, affine transformations provide better accuracy recovery against all adversarial attacks, with the correct classification for the FGSM attack on ResNet-50 and a significant top-3 accuracy recovery of 90.3\% for DenseNet-121. Finally, the defensive effectiveness of affine transformations against adversarial attacks remains an open subject, as the granular effect of different affine transformations has not been assessed, and no gray-box or black-box attacks were involved in the experiments.

Sun et al. \cite{sun2024affineattack} introduce an affine-invariant framework for enhanced adversarial attacks that apply to the face recognition task. Their generalized attack algorithm presents a broad set of results that favor their methodology over common attacks such as FGSM and PGD. At the same time, their approach also features performance improvements for query-based and transfer-based attacks. 

Finally, recent research~\cite{sandru2022fla} has enhanced the robustness of CNNs by applying affine transformations to the activation maps of intermediate layers, not just to the input data, thereby demonstrating their multi-scenario utility.

\subsection{Targeted Attacks}

Taori et al. \cite{taori2019targetedgenetic} introduce a black-box genetic algorithm approach for targeted attacks on audio systems. They add noise to an audio sample within a genetic algorithm framework. Hence, the output decoding scope is similar to a target, but the audio content is almost identical to the initial example. The method achieved up to 35\% of the input data that matched the target, with a 94.6\% similarity score for the output compared to the input. 

Another approach for generating targeted attacks is demonstrated by Kwon et al. \cite{kwon2018multitargetadversarial}, which involves obtaining perturbed data on multiple targets using a transformer that adds noise to both the sample and target models, thereby generating a loss. Their method aims to maximize the attack success of an adversarial example for a model across all target models. The results show a 100\% targeted attack success rate in the MNIST \cite{lecun2010mnist} benchmark dataset. 

Byun et al. \cite{byun2022transfertargetedattack} tested cross-model targeted attacks by generating images using projections on 3D objects and transforming them into 2D adversarial examples using a differentiable renderer. Cross-model experiments involve tuning the adversarial attack based on the loss of the source model, and the resulting images are then passed to a target model; thus, the target is the model that explores the transferability property of adversarial examples \cite{demontis2019transferableadversarial,waseda2021transferableadversarial}. Their object-based diverse input (ODI) method yields an 18.7\% improvement in the average targeted adversarial attack compared to the state of the art.

Another technique of crafting and performing targeted adversarial attacks is demonstrated by Di Noia et al. \cite{dinoia2020targetedmultimedia}, and it is also a method more similar to ours in terms of targets. The datasets are based on images of real-world fashion recommender systems. By minimizing the distance between an input sample and an adversarial output based on white-box adversarial networks such that the model classifies it as a target class, the authors provide valuable information on altering origin-to-target labels, with up to 100\% attack success for specific source class-to-target class experiments.

\begin{figure*}[htb!]
  \centering
  \includegraphics[width=1.025\textwidth]{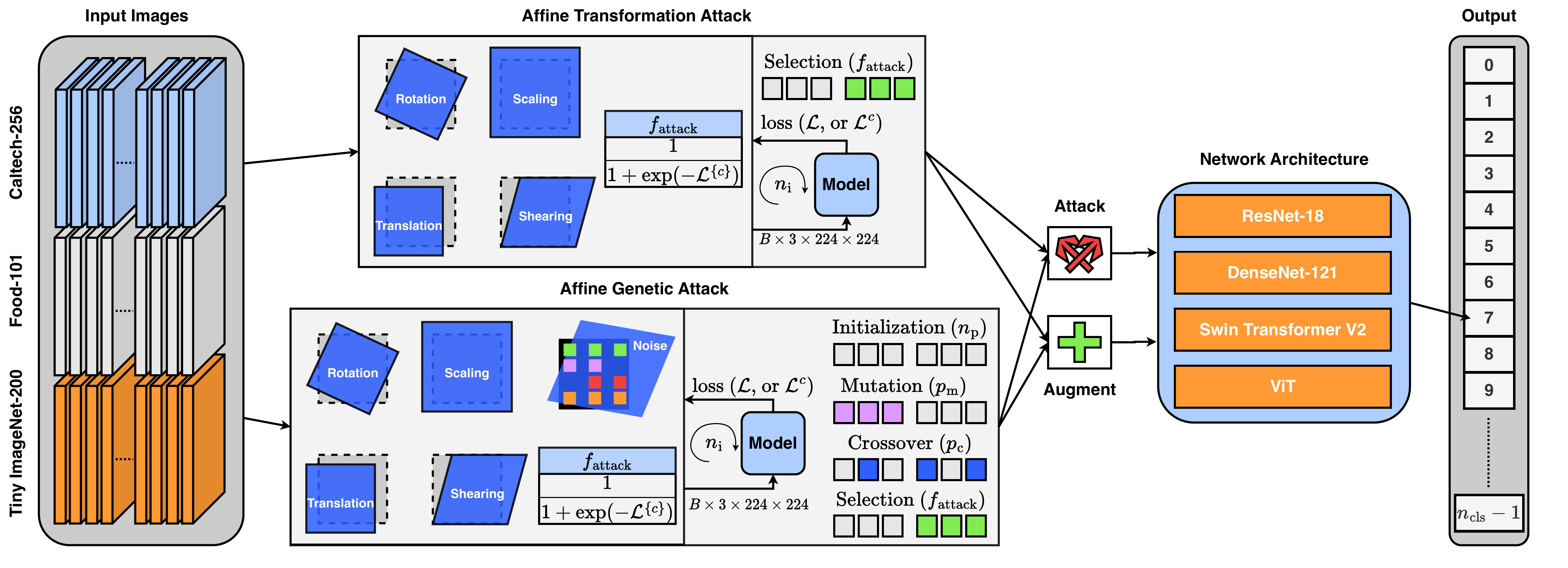}
  \caption{Affine Transformation Attack (ATA) and Affine Genetic Attack (AGA) algorithms. We add batches of images to the adversarial algorithms and obtain mutated images, which we then pass to the network architectures in either the training phase (Augment) or the testing phase (Attack) to achieve better performance or defense against attacks. We use the same selection method for both algorithms, maximizing $f_\text{attack}$ over several iterations ($n_i$). The AGA algorithm involves additional genetic stages: initialization ($n_p$, the population size), mutation with the probability $p_m$, and crossover with the probability $p_c$.}
   \label{fig:ata-aga-method}
\end{figure*}


\section{Methodology}

\subsection{Dataset}

The \textbf{Tiny-ImageNet-200} dataset is a subset of the larger ImageNet benchmark~\cite{russakovsky2015imagenet} and comprises $100{,}000$ training images and $10{,}000$ validation images spread uniformly across $200$ different classes (a total of $550$ images per class). The public dataset also contains $10{,}000$ unlabeled test images, which we did not use in our experiments. All images have a resolution of $64\times64$ pixels and correspond to nouns from the WordNet hierarchy~\cite{miller-1994-wordnet}.
The \textbf{Caltech-256} dataset~\cite{griffin2022caltech} contains $30{,}607$ images of varying sizes and aspect ratios. The images represent $257$ object categories and are unevenly distributed in these classes, with some categories being overrepresented. 
The \textbf{Food-101} dataset~\cite{bossard2014food} contains a total of $101{,}000$ images of popular dishes. They are spread evenly over $101$ classes, with each class assigned $750$ training images and $250$ test images of size $512 \times 512$.

\begin{figure}[htb!]
    \centering
    \includegraphics[width=1.01\columnwidth]{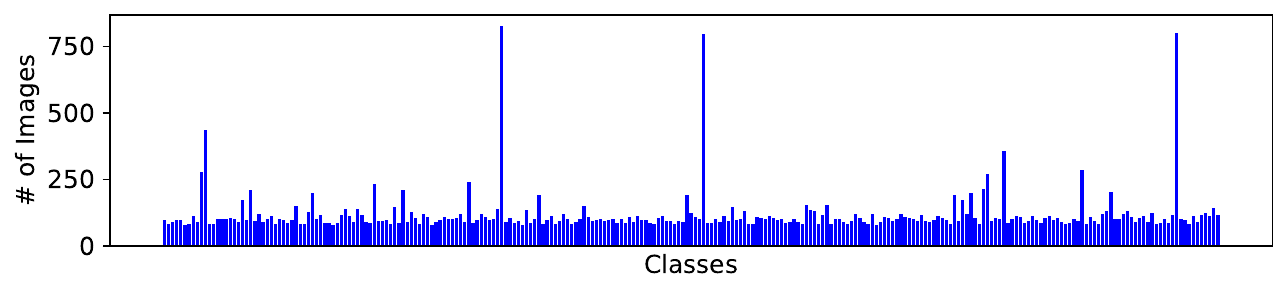}
    \caption{Caltech-256 class distribution. The classes are sorted alphabetically.}
    \label{fig:anx-dataset-distributions}
\end{figure}

 The only dataset with a variable class distribution used in our experiments, as shown in \cref{fig:anx-dataset-distributions}, is the Caltech-256 dataset, with a minimum class support of 80 images per class and a maximum of 827 images per class. The Tiny-ImageNet-200 and Food-101 datasets have constant support, with 550 images per class for Tiny-ImageNet-200 and 1000 images per class for Food-101, respectively. In the context of the different dataset characteristics, we define the dataset diversity factor ($d_f$) as the number of images per class, on average, normalized over the total number of classes:

\begin{equation}
\label{eq:a.1} 
d_f = \frac{\overline{n_\text{img}}}{n_{\text{cls}}}
\end{equation}
, where $\overline{n_\text{img}}$ is the average number of images per class for the entire dataset, and $n_{\text{cls}}$ is the number of classes for the dataset.

Considering the dataset diversity factor defined in \cref{eq:a.1}, we underline our choice for these datasets because of their contrasts in size and classes, where $d_f=0.46$ for Caltech-256, $2.75$ for Tiny-ImageNet-200, and $9.90$ for Food-101. We apply \cref{eq:a.1} to obtain balanced difficulty insights, that is, how the experimental results vary between dense versus underrepresented datasets.

We process the datasets for the experiments in this paper by resizing the images to $224 \times 224$ pixels. We divide each set into train/validation/test subsets with a 0.8/0.1/0.1 proportion.

\subsection{Model Architectures}

For our experiments, we test on two neural networks based on convolutional layers, \textbf{ResNet-18}~\cite{he2016resnet} and \textbf{DenseNet-121}~\cite{huang2017densenet}. For the DenseNet-121 network architecture, we set a dropout rate of 0.4.

The second type of model we evaluate is based on the attention mechanism and the transformer architecture~\cite{NIPS2017_3f5ee243}. \textbf{Vision Transformer (ViT)}~\cite{dosovitskiy2021vit} splits an image into disjoint patches of size $16\times16$ and transforms them into a sequence of ``tokens'', which is processed the same way as for natural language tasks. This renowned architecture has achieved state-of-the-art scores in numerous studies, usually pre-trained on large datasets and fine-tuned for downstream tasks. 
\textbf{Swin Transformer V2}~\cite{liu2022swinV2} is a more GPU memory-efficient architecture, compared to the previous version~\cite{liu2022swin}, which increases the model capacity and the window scale of the hierarchical transformer.
We use the SwinV2-T flavor of the architecture for our experiments, configured with $C=96$ (arbitrary projection dimension in the Swin Transformer architecture) and the number of blocks \{2, 2, 6, 2\}.

\subsection{Adversarial Algorithms}

As shown in \cref{fig:ata-aga-method}, the attack score is computed as follows:
\begin{equation}
\label{eq:1.1} 
f_{\text{attack}} =
\begin{cases}
\frac{1}{1+\exp(-\mathcal{L}^{c})}, & \text{if targeted attack.} \\
\frac{1}{1+\exp(-\mathcal{L})}, & \text{otherwise.}
\end{cases}
\end{equation}
, where $\mathcal{L}$ is the loss resulting from the validation of the images passed to the model, and $\mathcal{L}^c$ is the loss validating the images fed to the model against a target class $c\in[0,n_{\text{cls}}-1]$.

The \Cref{alg:ATA} describes our Affine Transformation Attack approach, and \Cref{alg:AGA} describes our Affine Genetic Attack method. We denote $\mathrm{CE}$ as Cross-Entropy loss, $\theta$ as rotation angle, $\tau_{x,y}$ as translation angle, $s$ as scaling factor, and $\phi$ as shearing angle. In our calculations, detailed in \cref{alg:ATA} and \cref{alg:AGA}, we introduce $\mathcal{S}$ and $\mathcal{S}^\ast$ to store the attack scores. Moreover, in \cref{alg:AGA}, $\Delta$ stores the sampled noise. The input values used in our experiments are: number of algorithm iterations $n_i=7$, population size $n_p=3$, mutation probability $p_m=0.3$, crossover probability $p_c=0.3$, adversarial intensity $\epsilon=0.1$. For brevity, individual image operations, such as applying affine transformations and mutations to particular images, are not detailed in \cref{alg:AGA}.

\begin{algorithm}[ht]
\caption{\small Affine Transformation Attack (ATA)}
\small
\label{alg:ATA}
\SetKwInOut{Input}{Input}\SetKwInOut{Output}{Output}

\Input{Batch $X\in[0,1]^{B\times3\times224\times224}$, Labels $y$, Model $\mathcal{M}$, Iterations $n_i$, (Optional) Target class $c$}
\Output{Adversarial images $\hat X^\ast$}

\BlankLine

$\hat X^\ast\leftarrow X.\mathrm{clone}()$\;

$\tilde y\leftarrow$ ($c$ if given else $y$)\;

$\mathcal{S}^\ast\leftarrow f_{\text{attack}}\big(\mathrm{CE}(\mathcal{M}(\hat X^\ast),\ \tilde y)\big)$\;

\For{$t=0\ldots n_i-1$}{
    $\widetilde{X}\leftarrow X.\mathrm{clone}()$\;
    
    \For{$j=0\ldots B-1$}{
        Sample $\theta\sim\mathcal{U}(-3,3), \tau_{x,y}\sim\mathcal{U}(-0.05,0.05),$ \\ 
                $\ \ \ \ \ \ \ \ \ \ \ \ \ s\sim\mathcal{U}(0.95,1.05), \phi\sim\mathcal{U}(-1,1)$\;
        Apply affine transformations to $\widetilde{X}[j]$ with $\theta,\tau_{x,y},s,\phi$;
    }

    $\mathcal{L}^{\{c\}}\leftarrow\mathrm{CE}(\mathcal{M}(\widetilde{X}),\tilde y)$\; 
    $\mathcal{S}\leftarrow f_{\text{attack}}(\mathcal{L}^{\{c\}})$;

    \If{$\mathcal{S}>\mathcal{S}^\ast$}{
        $\mathcal{S}^\ast\leftarrow\mathcal{S}$\; 
        $\hat X^\ast\leftarrow\widetilde{X}$\;
    }
}
\Return $\hat X^\ast$\;
\end{algorithm}

\textbf{Affine Transformation Attack (ATA).} 
Inspired by Athalye et al. \cite{athalye2018affine}, Shen et al. \cite{shen2019affine}, and Sandru et al. \cite{sandru2022fla}, our first iterative algorithm, ATA, includes affine transformations (rotation, translation, scaling, shearing) generated based on random uniform samples as described in \cref{alg:ATA}. In contrast to the Sandru et al. \cite{sandru2022fla} method, we narrow the ranges using finely transformed examples to avoid image alteration by iterative operations.

We pass the adversarial batch to the model, and compute the attack score $f_{\text{attack}}$ based on \cref{eq:1.1}. After all iterations, the algorithm returns the most suitable adversarial images (the \textbf{selection} phase). For untargeted attacks, the algorithm maximizes the score based on the ground truth, whereas, in the targeted adversarial configuration, it maximizes the score based on the smallest Cross-Entropy loss ($\mathcal{L}^{\{c\}}$) for every picture being classified as the target class.

\begin{algorithm}[ht]
\caption{\small Affine Genetic Attack (AGA)}
\small
\label{alg:AGA}
\SetKwInOut{Input}{Input}\SetKwInOut{Output}{Output}

\Input{Batch $X\in[0,1]^{B\times3\times224\times224}$, Labels $y$, Model $\mathcal{M}$, Iterations $n_i$, Population $n_p$, Mutation $p_m$, Crossover $p_c$, Adversarial intensity $\varepsilon$, (Optional) Target class $c$}
\Output{Adversarial images $\hat X^\ast$}

\BlankLine

$\mathcal{P}\leftarrow\mathrm{repeat}(X,n_p)$\;
$\tilde y\leftarrow$ ($c$ if given else $y$)\;

\For{$t=0\ldots n_i-1$}{
    $\widetilde{\mathcal{P}}\leftarrow\mathcal{P}.\mathrm{clone}()$\;
    
    \For{$j=0\ldots n_p-1$}{
        \If{$p\sim\mathcal{U}(0,1)<p_m$}{ 
            Sample $\theta\sim\mathcal{U}(-3,3), \tau_{x,y}\sim\mathcal{U}(-0.05,0.05),$ \\ 
                    $\ \ \ \ \ \ \ \ \ \ \ \ \ s\sim\mathcal{U}(0.95,1.05), \phi\sim\mathcal{U}(-1,1)$\;
            Draw $\Delta\sim\mathcal{U}(0,\varepsilon)^{B\times3\times224\times224}$\; 
            Apply affine transformations to $\widetilde{\mathcal{P}}[j]$ with $\theta, \tau_{x,y}, s, \phi$\;
            $\widetilde{\mathcal{P}}[j]\leftarrow\operatorname{clamp}(\widetilde{\mathcal{P}}[j]+\Delta,0,1)$\;
        }
    }
    
    
    \For{$j=0,2,\dots,n_p-2$}{ 
        \If{$p\sim\mathcal{U}(0,1)<p_c$}{ 
            Sample $r\in\{1,\dots,223\}$\;
            Swap rows $0{:}r$ between $\widetilde{\mathcal{P}}[j]$ and $\widetilde{\mathcal{P}}[j+1]$\;
        } 
    }
        
    \For{$j=0\ldots n_p-1$}{ 
        $\mathcal{L}^{\{c\}}_j\leftarrow\mathrm{CE}(\mathcal{M}(\widetilde{\mathcal{P}}[j]),\tilde y)$\; 
        $\mathcal{S}_j\leftarrow{f_{\text{attack}}}(\mathcal{L}^{\{c\}}_j)$\;
    }
    
    $k\leftarrow\arg\max_j \mathcal{S}_j$\;
    $\tilde X^\ast\leftarrow\widetilde{\mathcal{P}}[k]$\;
    $\mathcal{P}\leftarrow\mathrm{repeat}(\tilde X^\ast,n_p)$;
}
\Return $\hat X^\ast \leftarrow \big(\dim(\tilde X^\ast)=5\big)\ ?\ \tilde X^\ast[0]\ :\ \tilde X^\ast$\;
\end{algorithm}

\textbf{Affine Genetic Attack (AGA).}
In addition to a simple iterative affine-only attack, AGA augments adversarial exploration with genetic operators and random noise. 
In \cref{alg:AGA}, the algorithm starts with the \textbf{initialization} phase by cloning the input images into a population of candidates. During \textbf{mutation}, each individual, with probability $p_m$, is perturbed by random affine transformations and bounded pixel noise scaled by an adversarial intensity $\varepsilon$. A \textbf{crossover} phase follows, where neighboring candidates exchange image segments, with probability $p_c$, creating recombined offspring. In the \textbf{selection} step, all candidates are scored via the model loss ($\mathcal{L}^{\{c\}}$) mapped to the attack score $f_{\text{attack}}$ (see \cref{eq:1.1}). The highest-scoring individual is chosen and replicated to reinitialize the population for the next generation. After a fixed number of iterations, through ground-truth evaluation or targeting a specific class, the algorithm returns the best adversarial images, that is, the fittest batch from the resulting population.

\subsection{Experimental Setup}

\textbf{Data augmentation.}
We evaluate the performance with and without data augmentations for each dataset and model architecture. By doubling the training set while generating additional adversarial examples, we obtain an extra loss used in the model fine-tuning.

\textbf{Adversarial attacks.}
We refer to an untargeted attack as a general mutation of the entire set passed to the algorithm, without focusing on improving performance in a specific class of the dataset. In this configuration, we use the accuracy and the attack success rate (\cref{eq:1.2}). We evaluate the model's capability for undefended attacks, with no adversarial data augmentation used in training, and defended ones, using augmentations generated by the attack algorithms. Unlike the untargeted attack, the targeted attack aims to shift the classification performance towards a specific class. As the dataset is unbalanced, we utilize Caltech-256 to assess the targeted attack capability and evaluate each model architecture in our algorithms.

\textbf{Algorithm parameter variation.}
For the models trained without augmentations, we vary the parameters of the attack algorithm for each dataset to assess the sensitivity of the parameters and the success of the attack of each configuration. 
For both ATA and AGA, we vary the number of black-box iterations ($n_i$) in the range [1, 10], and, for AGA only, we also iterate over $p_c$, $p_m$, and $\epsilon$ in the range [0, 1].

\textbf{Training hyperparameters and optimization.}
The selected model architectures are trained for 12 epochs using a batch size of 32, an initial learning rate of $ 10^ {- 4} $ with linear decay and two warm-up epochs, and the Adam optimizer \cite{kingma2014adam}.

\textbf{Evaluation metrics.} 
The main metrics used in our work are validation and test accuracy. We also include the macro and weighted F1 scores for a better global view. For adversarial experiments, we define the following attack success rate (SR) score:
\begin{equation}
\label{eq:1.2} 
SR = \left( 1 - \frac{Acc_{\text{attacked}}} {Acc_{\text{unattacked}}} \right)  * 100 
\end{equation}
, where $Acc_{\text{unattacked}}$ represents the accuracy of testing the model with the initial test set, and $Acc_{\text{attacked}}$ represents the accuracy resulting from the evaluation of the attacked (mutated) version of the test set.

\textbf{Statistical stability.}
As our algorithms utilize random samples for adversarial data generation, we run our setup through five iterations for each set of experiments and present our results as the mean and standard deviation. The exception is for targeted attack experiments, where the results are illustrated using the average attack success rate without the standard deviation.

\textbf{Experimental environment and limitations.} 
For our experiments, we used a Tesla A100 40GB GPU. The primary limitation of our algorithms, notably the AGA algorithm, is the hardware's memory capacity. Given that the genetic population size is a tunable parameter, we limited the population size ($n_i$) to 3, considering our hardware setup. We also restrict the batch size to 32 to facilitate training with adversarial examples in the ViT architecture.
Moreover, given the size of the datasets, training with adversarial data obtained iteratively will also result in a significant increase in the time required, proportional to the number of iterations ($n_i$).


\section{Results}

\subsection{Model Performance}

\begin{table}[htb]

\caption{Model training results. We assess our adversarial algorithms, Affine Transform Attack (ATA) and Affine Genetic Attack (AGA), on data augmentation and compare with Sandru et al. \cite{sandru2022fla} baseline (Base) and Feature-Level Augmentation (FLA), and with our method without data augmentation (No Aug.). Performance is evaluated through validation and test accuracy (V. Acc. and T. Acc.), as well as macro and weighted F1 scores (M. F1 and W. F1). The ResNet-18 (RN-18), DenseNet-121 (DN-121), and Swin Transformer V2 (STV2) architectures are abbreviated for spacing reasons. The $\uparrow$ indicates that higher values represent better classification performance.}
\centering

\resizebox{\columnwidth}{!}{
\begin{tabular}{|c|c|l|l|l|l|l|}
\hline
\textbf{Dataset}                    & \textbf{Arch.}            & \multicolumn{1}{c|}{\textbf{Method}}         & \multicolumn{1}{c|}{\textbf{V. Acc. (\%) $\uparrow$}} & \multicolumn{1}{c|}{\textbf{T. Acc. (\%) $\uparrow$}} & \multicolumn{1}{c|}{\textbf{M. F1 (\%) $\uparrow$}} & \multicolumn{1}{c|}{\textbf{W. F1 (\%) $\uparrow$}} \\ \hline \hline
\multirow{16}{*}{\rotatebox[origin=c]{90}{Caltech-256}}       & \multirow{5}{*}{\rotatebox[origin=c]{90}{RN-18}}        & AGA                                             & 78.23 $\pm$ 0.25                                        & 78.02 $\pm$ 0.12                                  & 80.08  $\pm$ 0.08                                     & 78.29 $\pm$ 0.42                                         \\ \cline{3-7} 
                                    &                                  & ATA                                             & \textbf{81.57 $\pm$ 0.18}                                 & \textbf{80.77 $\pm$ 0.17}                          & \textbf{82.24 $\pm$ 0.25}                              & \textbf{80.57 $\pm$ 0.22}                                 \\ \cline{3-7} 
                                    &                                  & No Aug.                                 & 80.76 $\pm$ 0.38                                        & 80.02 $\pm$ 0.17                                  & 81.57 $\pm$ 0.24                                      & 79.76 $\pm$ 0.19                                        \\ \cline{3-7} 
                                    &                                  & Base \cite{sandru2022fla} & -                                              & 78.96 $\pm$ 0.30           & -                                           & -                                              \\ \cline{3-7} 
                                    &                                  & FLA \cite{sandru2022fla}      & -                                              & 78.91 $\pm$ 0.06           & -                                           & -                                              \\ \cline{2-7} 
                                    & \multirow{5}{*}{\rotatebox[origin=c]{90}{DN-121}}     & AGA                                             & 83.97 $\pm$ 0.20                                         & 83.80 $\pm$ 0.06                                  & 84.84  $\pm$ 0.15                                     & 83.60 $\pm$ 0.04                                         \\ \cline{3-7} 
                                    &                                  & ATA                                             & 85.32 $\pm$ 0.05                                         & 84.41 $\pm$ 0.15                                  & 84.96  $\pm$ 0.18                                     & 84.05 $\pm$ 0.15                                         \\ \cline{3-7} 
                                    &                                  & No Aug.                                 & \textbf{85.89 $\pm$ 0.15}                                 & \textbf{84.47 $\pm$ 0.12}                          & \textbf{85.58 $\pm$ 0.22}                              & \textbf{84.28 $\pm$ 0.14}                                 \\ \cline{3-7} 
                                    &                                  & Base \cite{sandru2022fla} & -                                              & 83.98 $\pm$ 0.14           & -                                           & -                                              \\ \cline{3-7} 
                                    &                                  & FLA \cite{sandru2022fla}      & -                                              & 83.54 $\pm$ 0.27           & -                                           & -                                              \\ \cline{2-7} 
                                    & \multirow{3}{*}{\rotatebox[origin=c]{90}{STV2}} & AGA                                             & 90.25 $\pm$ 0.11                                          & 89.73 $\pm$ 0.41                                   & 90.43 $\pm$ 0.26                                       & 89.67 $\pm$ 0.29                                          \\ \cline{3-7} 
                                    &                                  & ATA                                             & \textbf{90.69 $\pm$ 0.17}                                 & 89.53 $\pm$ 0.15                                  & 90.11 $\pm$ 0.10                                       & 89.51 $\pm$ 0.14                                          \\ \cline{3-7} 
                                    &                                  & No Aug.                                 & 90.62 $\pm$ 0.19                                          & \textbf{90.16 $\pm$ 0.16}                          & \textbf{90.74 $\pm$ 0.11}                              & \textbf{90.06 $\pm$ 0.17}                                 \\ \cline{2-7} 
                                    & \multirow{3}{*}{\rotatebox[origin=c]{90}{ViT}}             & AGA                                             & 89.30 $\pm$ 0.15                                          & 89.17 $\pm$ 0.09                                   & 88.47 $\pm$ 0.05                                       & 89.19 $\pm$ 0.14                                          \\ \cline{3-7} 
                                    &                                  & ATA                                             & 89.13 $\pm$ 0.25                                          & \textbf{89.72 $\pm$ 0.12}                          & \textbf{88.80 $\pm$ 0.08}                              & \textbf{89.68 $\pm$ 0.17}                                 \\ \cline{3-7} 
                                    &                                  & No Aug.                                 & \textbf{89.50 $\pm$ 0.29}                                 & 89.41 $\pm$ 0.18                                   & 88.70 $\pm$ 0.11                                       & 89.45 $\pm$ 0.21                                          \\ \hline \hline
\multirow{16}{*}{\rotatebox[origin=c]{90}{Food-101}}          & \multirow{5}{*}{\rotatebox[origin=c]{90}{RN-18}}        & AGA                                             & 72.82  $\pm$ 0.13                                        & 73.25 $\pm$ 0.19                                   & \textbf{73.12 $\pm$ 0.16}                              & \textbf{73.12 $\pm$ 0.09}                                 \\ \cline{3-7} 
                                    &                                  & ATA                                             & \textbf{73.74 $\pm$ 0.18}                                 & 72.31 $\pm$ 0.29                                   & 72.09 $\pm$ 0.12                                       & 72.09 $\pm$ 0.19                                          \\ \cline{3-7} 
                                    &                                  & No Aug.                                 & 73.37 $\pm$ 0.21                                          & 72.99 $\pm$ 0.05                                   & 72.78 $\pm$ 0.08                                       & 72.78 $\pm$ 0.03                                          \\ \cline{3-7} 
                                    &                                  & Base \cite{sandru2022fla} & -                                              & \textbf{76.29 $\pm$ 0.41}  & -                                           & -                                              \\ \cline{3-7} 
                                    &                                  & FLA \cite{sandru2022fla}      & -                                              & 76.28 $\pm$ 0.33           & -                                           & -                                              \\ \cline{2-7} 
                                    & \multirow{5}{*}{\rotatebox[origin=c]{90}{DN-121}}     & AGA                                             & 79.97 $\pm$ 0.18                                          & 79.32 $\pm$ 0.21                                   & \textbf{79.19 $\pm$ 0.22}                              & \textbf{79.19 $\pm$ 0.12}                                 \\ \cline{3-7} 
                                    &                                  & ATA                                             & \textbf{80.09 $\pm$ 0.17}                                 & 79.42 $\pm$ 0.12                                   & 79.16 $\pm$ 0.20                                       & 79.16  $\pm$ 0.12                                         \\ \cline{3-7} 
                                    &                                  & No Aug.                                 & 79.34 $\pm$ 0.11                                          & 78.59 $\pm$ 0.14                                   & 78.42 $\pm$ 0.09                                       & 78.42 $\pm$ 0.06                                          \\ \cline{3-7} 
                                    &                                  & Base \cite{sandru2022fla} & -                                              & \textbf{83.26 $\pm$ 0.20}  & -                                           & -                                              \\ \cline{3-7} 
                                    &                                  & FLA \cite{sandru2022fla}      & -                                              & 82.86 $\pm$ 0.20           & -                                           & -                                              \\ \cline{2-7} 
                                    & \multirow{3}{*}{\rotatebox[origin=c]{90}{STV2}} & AGA                                             & 85.08 $\pm$ 0.19                                          & 84.32 $\pm$ 0.22                                   & 84.26 $\pm$ 0.16                                       & 84.25 $\pm$ 0.27                                          \\ \cline{3-7} 
                                    &                                  & ATA                                             & \textbf{85.79 $\pm$ 0.14}                                 & \textbf{84.88 $\pm$ 0.12}                          & \textbf{84.77 $\pm$ 0.07}                              & \textbf{84.76 $\pm$ 0.14}                                 \\ \cline{3-7} 
                                    &                                  & No Aug.                                 & 83.74 $\pm$ 0.18                                          & 82.70 $\pm$ 0.22                                   & 82.66 $\pm$ 0.08                                       & 82.65 $\pm$ 0.15                                          \\ \cline{2-7} 
                                    & \multirow{3}{*}{\rotatebox[origin=c]{90}{ViT}}             & AGA                                             & 82.60 $\pm$ 0.18                                          & 81.90  $\pm$ 0.32                                  & 81.91 $\pm$ 0.17                                       & 81.91  $\pm$ 0.24                                          \\ \cline{3-7} 
                                    &                                  & ATA                                             & \textbf{84.12 $\pm$ 0.18}                                 & \textbf{83.66 $\pm$ 0.34}                          & \textbf{83.65 $\pm$ 0.11}                              & \textbf{83.65 $\pm$ 0.23}                                 \\ \cline{3-7} 
                                    &                                  & No Aug.                                 & 81.77 $\pm$ 0.14                                          & 82.14 $\pm$ 0.16                                   & 82.13 $\pm$ 0.11                                       & 82.13 $\pm$ 0.09                                         \\ \hline \hline
\multirow{16}{*}{\rotatebox[origin=c]{90}{Tiny-ImageNet-200}} & \multirow{5}{*}{\rotatebox[origin=c]{90}{RN-18}}        & AGA                                             & 69.85 $\pm$ 0.32                                          & 69.65 $\pm$ 0.29                                  & 69.28 $\pm$ 0.15                                       & 69.25 $\pm$ 0.28                                          \\ \cline{3-7} 
                                    &                                  & ATA                                             & \textbf{71.94 $\pm$ 0.17}                                 & 71.23 $\pm$ 0.09                                   & \textbf{70.87 $\pm$ 0.16}                              & \textbf{70.83 $\pm$ 0.23}                                 \\ \cline{3-7} 
                                    &                                  & No Aug.                                 & 70.93 $\pm$ 0.05                                          & 70.96 $\pm$ 0.18                                   & 70.73 $\pm$ 0.11                                       & 70.72 $\pm$ 0.21                                          \\ \cline{3-7} 
                                    &                                  & Base \cite{sandru2022fla} & -                                              & 71.50 $\pm$ 0.20           & -                                           & -                                              \\ \cline{3-7} 
                                    &                                  & FLA \cite{sandru2022fla}      & -                                              & \textbf{71.76 $\pm$ 0.16}  & -                                           & -                                              \\ \cline{2-7} 
                                    & \multirow{5}{*}{\rotatebox[origin=c]{90}{DN-121}}     & AGA                                             & 75.28 $\pm$ 0.24                                          & 74.55 $\pm$ 0.29                                   & 74.08 $\pm$ 0.18                                       & 74.08 $\pm$ 0.15                                          \\ \cline{3-7} 
                                    &                                  & ATA                                             & \textbf{76.96 $\pm$ 0.11}                                 & 75.34 $\pm$ 0.14                                   & \textbf{74.83 $\pm$ 0.16}                              & \textbf{74.83 $\pm$ 0.12}                                 \\ \cline{3-7} 
                                    &                                  & No Aug.                                 & 76.57 $\pm$ 0.15                                          & 75.20 $\pm$ 0.38                                   & 74.77 $\pm$ 0.17                                       & 74.77 $\pm$ 0.13                                          \\ \cline{3-7} 
                                    &                                  & Base \cite{sandru2022fla} & -                                              & 76.50 $\pm$ 0.37           & -                                           & -                                              \\ \cline{3-7} 
                                    &                                  & FLA \cite{sandru2022fla}      & -                                              & \textbf{76.60 $\pm$ 0.44}  & -                                           & -                                              \\ \cline{2-7} 
                                    & \multirow{3}{*}{\rotatebox[origin=c]{90}{STV2}} & AGA                                             & 82.51 $\pm$ 0.26                                          & 81.93 $\pm$ 0.19                                   & 81.85 $\pm$ 0.11                                       & 81.85 $\pm$ 0.21                                          \\ \cline{3-7} 
                                    &                                  & ATA                                             & 83.38 $\pm$ 0.35                                        & 83.16 $\pm$ 0.11                                  & 83.10 $\pm$ 0.15                                      & 83.10 $\pm$ 0.14                                          \\ \cline{3-7} 
                                    &                                  & No Aug.                                 & \textbf{84.02 $\pm$ 0.29}                                 & \textbf{83.55 $\pm$ 0.26}                          & \textbf{83.47 $\pm$ 0.16}                              & \textbf{83.47 $\pm$ 0.24}                                 \\ \cline{2-7} 
                                    & \multirow{3}{*}{\rotatebox[origin=c]{90}{ViT}}             & AGA                                             & 83.99 $\pm$ 0.15                                          & 84.01 $\pm$ 0.12                                   & 83.99 $\pm$ 0.08                                       & 83.99 $\pm$ 0.11                                          \\ \cline{3-7} 
                                    &                                  & ATA                                             & 85.22 $\pm$ 0.24                                          & 84.66 $\pm$ 0.19                                   & 84.66 $\pm$ 0.14                                       & 84.66 $\pm$ 0.15                                          \\ \cline{3-7} 
                                    &                                  & No Aug.                                 & \textbf{85.95 $\pm$ 0.25}                                 & \textbf{85.42 $\pm$ 0.31}                          & \textbf{85.41 $\pm$ 0.29}                              & \textbf{85.41 $\pm$ 0.35}                                 \\ \hline
\end{tabular}

}

\label{tab:results-model-performance}
\end{table}

\textbf{Dataset comparison.}
Based on the results of \cref{tab:results-model-performance}, we obtain the best classification accuracy on the Caltech-256 dataset, with an average of 90.16\%, followed by Tiny-ImageNet-200 with 85.42\%, and Food-101 with 84.88\%. On the F1 score, we attain the best results on the Caltech-256 dataset, with a macro score of 90.74\% and a weighted score of 90.06\%. Therefore, we achieve 5.28\% better accuracy on Caltech-256 compared to the Food-101 dataset and 4.74\% increased accuracy on Caltech-256 compared to Tiny-ImageNet-200.

\textbf{Model architecture comparison.}
Comparing the results of various model architectures, as shown in \cref{tab:results-model-performance}, we reach better results with computer vision transformer architectures. The best results are achieved with the Swin Transformer V2 models for the Caltech-256 and Food-101 datasets (10.14\% and 12.57\%, representing added value compared to the same augmentation algorithm), as well as ViT for Tiny-ImageNet-200, yielding a 14.46\% accuracy gain.

\textbf{Algorithm comparison.}
We compare the data augmentation results of our adversarial algorithms, depicted in \cref{tab:results-model-performance}, to the Sandru et al. \cite{sandru2022fla} baseline and Feature-Level Augmentation (FLA). 
Testing on ResNet-18 and DenseNet-121, Sandru et al. \cite{sandru2022fla} provide better results, with 76.29\% and 83.26\% accuracy on the Food-101 dataset (their baseline) and 71.76\% and 76.60\% on the Tiny-ImageNet-200 (FLA). For Caltech-256, we outperform their method with an average accuracy of 80.77\% (+1.91\% compared to their best results) on ResNet-18 (ATA algorithm) and 84.47\% (+0.49\%) on DenseNet-121 without augmentation. For the computer vision transformer experiments, we outperform all the data augmentation results of Sandru et al. \cite{sandru2022fla}. For the Caltech-256 and Food-101 datasets, our best scores are achieved with Swin Transformer V2 on training without data augmentations, attaining an average accuracy of 90.16\% (+6.18\% compared to Sandru et al. \cite{sandru2022fla} best results), and 84.88\% using ATA augmentations (+1.62\%). For Tiny-ImageNet-200, the ViT architecture performs better than Swin Transformer V2, with an average accuracy of 85.42\% using no data augmentation (+8.82\% compared to Sandru et al. \cite{sandru2022fla} and +1.87\% over the Swin Transformer V2).

\textbf{Consistency and stability of the results.} For macro and weighted F1 scores, we reflect consistency with the classification accuracy, with 90.74\% macro F1 and 90.06\% weighted F1 for our best results on Caltech-256, 84.77\% and 84.76\% on Food-101, and 85.41\% and 85.41\% on Tiny-ImageNet-200. Regarding the stability of the test results over multiple training iterations, we observe a standard deviation of no more than 0.5\% on all our metrics (validation accuracy, test accuracy, macro F1, and weighted F1).

\subsection{Adversarial Attacks}

\begin{table*}[!hbt]

\caption{Untargeted attack results. We evaluate Affine Transform Attack (ATA) and Affine Genetic Attack (AGA) algorithms for every dataset and model architecture on adversarial attacks for both defended (models trained with adversarial data augmentations) and undefended (no augmentation in training) scenarios and reflect performance through attack accuracy (Attack Acc.), the accuracy on the adversarial test set, and attack success rate (SR), where $\downarrow$ means better attacks for lower accuracy and $\uparrow$ represents a better attack for higher SR.}

\centering

\resizebox{2.079\columnwidth}{!}{
\begin{tabular}{|c|c|llll|llll|}
\hline
\multirow{3}{*}{\textbf{Dataset}}  & \multirow{3}{*}{\textbf{Architecture}} & \multicolumn{4}{c|}{\textbf{Attack Acc. (\%)   $\downarrow$}}                                                                                                                                              & \multicolumn{4}{c|}{\textbf{SR (\%) $\uparrow$}}                                                                                                                                                           \\ \cline{3-10} 
                                   &                                 & \multicolumn{2}{c|}{\textbf{AGA}}                                                                                & \multicolumn{2}{c|}{\textbf{ATA}}                                                       & \multicolumn{2}{c|}{\textbf{AGA}}                                                                                & \multicolumn{2}{c|}{\textbf{ATA}}                                                       \\ \cline{3-10} 
                                   &                                 & \multicolumn{1}{c|}{\textbf{No Aug.}}                       & \multicolumn{1}{c|}{\textbf{Aug.}}                 & \multicolumn{1}{c|}{\textbf{No Aug.}}              & \multicolumn{1}{c|}{\textbf{Aug.}} & \multicolumn{1}{c|}{\textbf{No Aug.}}                       & \multicolumn{1}{c|}{\textbf{Aug.}}                 & \multicolumn{1}{c|}{\textbf{No Aug.}}              & \multicolumn{1}{c|}{\textbf{Aug.}} \\ \hline \hline
\multirow{4}{*}{Caltech-256}       & ResNet-18                        & \multicolumn{1}{l|}{\textbf{46.85 $\pm$ 1.61}} & \multicolumn{1}{l|}{60.90 $\pm$ 0.74}  & \multicolumn{1}{l|}{69.86 $\pm$ 0.55} & 73.18 $\pm$ 0.35      & \multicolumn{1}{l|}{\textbf{41.45 $\pm$ 2.02}} & \multicolumn{1}{l|}{21.95 $\pm$ 0.95} & \multicolumn{1}{l|}{12.69 $\pm$ 0.68} & ~~9.40 $\pm$ 0.43        \\ \cline{2-10} 
                                   & DenseNet-121                     & \multicolumn{1}{l|}{\textbf{51.22 $\pm$ 0.90}}  & \multicolumn{1}{l|}{64.73 $\pm$ 0.41} & \multicolumn{1}{l|}{72.68 $\pm$ 0.51} & 75.80 $\pm$ 0.62       & \multicolumn{1}{l|}{\textbf{39.36 $\pm$ 1.06}} & \multicolumn{1}{l|}{22.75 $\pm$ 0.49} & \multicolumn{1}{l|}{13.96 $\pm$ 0.61} & 10.20 $\pm$ 0.73       \\ \cline{2-10} 
                                   & Swin Transformer V2                 & \multicolumn{1}{l|}{\textbf{51.30 $\pm$ 0.21}}  & \multicolumn{1}{l|}{74.77 $\pm$ 0.78} & \multicolumn{1}{l|}{79.62 $\pm$ 0.65} & 85.22 $\pm$ 0.46      & \multicolumn{1}{l|}{\textbf{43.10 $\pm$ 0.23}}  & \multicolumn{1}{l|}{16.67 $\pm$ 0.87} & \multicolumn{1}{l|}{11.69 $\pm$ 0.72} & ~~4.81 $\pm$ 0.51       \\ \cline{2-10} 
                                   & ViT                             & \multicolumn{1}{l|}{\textbf{62.39 $\pm$ 0.58}} & \multicolumn{1}{l|}{79.42 $\pm$ 0.60}  & \multicolumn{1}{l|}{86.01 $\pm$ 0.19} & 87.41 $\pm$ 0.13      & \multicolumn{1}{l|}{\textbf{30.22 $\pm$ 0.65}} & \multicolumn{1}{l|}{10.93 $\pm$ 0.68} & \multicolumn{1}{l|}{~~3.81 $\pm$ 0.21}  & ~~2.57 $\pm$ 0.14       \\ \hline \hline
\multirow{4}{*}{Food-101}          & ResNet-18                        & \multicolumn{1}{l|}{\textbf{32.68 $\pm$ 0.34}} & \multicolumn{1}{l|}{48.90 $\pm$ 0.31}  & \multicolumn{1}{l|}{60.08 $\pm$ 0.34} & 63.77 $\pm$ 0.45      & \multicolumn{1}{l|}{\textbf{55.22 $\pm$ 0.47}} & \multicolumn{1}{l|}{33.24 $\pm$ 0.42} & \multicolumn{1}{l|}{17.69 $\pm$ 0.46} & 11.80 $\pm$ 0.63       \\ \cline{2-10} 
                                   & DenseNet-121                     & \multicolumn{1}{l|}{\textbf{36.38 $\pm$ 0.60}}  & \multicolumn{1}{l|}{53.88 $\pm$ 0.34} & \multicolumn{1}{l|}{65.05 $\pm$ 0.31} & 69.33 $\pm$ 0.21      & \multicolumn{1}{l|}{\textbf{53.71 $\pm$ 0.76}} & \multicolumn{1}{l|}{32.07 $\pm$ 0.43} & \multicolumn{1}{l|}{17.23 $\pm$ 0.39} & 12.70 $\pm$ 0.27       \\ \cline{2-10} 
                                   & Swin Transformer V2                 & \multicolumn{1}{l|}{\textbf{36.31 $\pm$ 0.94}} & \multicolumn{1}{l|}{65.71 $\pm$ 0.32} & \multicolumn{1}{l|}{72.83 $\pm$ 0.31} & 79.05 $\pm$ 0.16      & \multicolumn{1}{l|}{\textbf{56.09 $\pm$ 1.14}} & \multicolumn{1}{l|}{22.06 $\pm$ 0.38} & \multicolumn{1}{l|}{11.94 $\pm$ 0.38} & ~~6.86 $\pm$ 0.19       \\ \cline{2-10} 
                                   & ViT                             & \multicolumn{1}{l|}{\textbf{41.21 $\pm$ 1.31}} & \multicolumn{1}{l|}{69.08 $\pm$ 0.18} & \multicolumn{1}{l|}{78.93 $\pm$ 0.21} & 81.71 $\pm$ 0.10       & \multicolumn{1}{l|}{\textbf{49.83 $\pm$ 1.59}} & \multicolumn{1}{l|}{15.65 $\pm$ 0.22} & \multicolumn{1}{l|}{~~3.91 $\pm$ 0.25}  & ~~2.34 $\pm$ 0.12       \\ \hline \hline
\multirow{4}{*}{Tiny-ImageNet-200} & ResNet-18                        & \multicolumn{1}{l|}{\textbf{27.83 $\pm$ 0.56}} & \multicolumn{1}{l|}{46.78 $\pm$ 0.27} & \multicolumn{1}{l|}{52.65 $\pm$ 0.26} & 60.72 $\pm$ 0.19      & \multicolumn{1}{l|}{\textbf{60.78 $\pm$ 0.78}} & \multicolumn{1}{l|}{32.83 $\pm$ 0.39} & \multicolumn{1}{l|}{25.80 $\pm$ 0.36}  & 14.75 $\pm$ 0.27      \\ \cline{2-10} 
                                   & DenseNet-121                     & \multicolumn{1}{l|}{\textbf{29.43 $\pm$ 0.67}} & \multicolumn{1}{l|}{48.95 $\pm$ 0.24} & \multicolumn{1}{l|}{57.03 $\pm$ 0.56} & 66.30 $\pm$ 0.32       & \multicolumn{1}{l|}{\textbf{60.86 $\pm$ 0.89}} & \multicolumn{1}{l|}{34.33 $\pm$ 0.32} & \multicolumn{1}{l|}{24.16 $\pm$ 0.74} & 11.99 $\pm$ 0.43      \\ \cline{2-10} 
                                   & Swin Transformer V2                 & \multicolumn{1}{l|}{\textbf{33.74 $\pm$ 0.58}} & \multicolumn{1}{l|}{65.44 $\pm$ 0.12} & \multicolumn{1}{l|}{74.50 $\pm$ 0.15}  & 76.42 $\pm$ 0.19      & \multicolumn{1}{l|}{\textbf{59.62 $\pm$ 0.69}} & \multicolumn{1}{l|}{20.12 $\pm$ 0.15} & \multicolumn{1}{l|}{10.83 $\pm$ 0.18} & ~~8.11 $\pm$ 0.23       \\ \cline{2-10} 
                                   & ViT                             & \multicolumn{1}{l|}{\textbf{30.77 $\pm$ 0.94}} & \multicolumn{1}{l|}{71.94 $\pm$ 0.18} & \multicolumn{1}{l|}{81.20 $\pm$ 0.11}  & 82.39 $\pm$ 0.10       & \multicolumn{1}{l|}{\textbf{63.97 $\pm$ 1.10}}  & \multicolumn{1}{l|}{14.37 $\pm$ 0.22} & \multicolumn{1}{l|}{~~4.94 $\pm$ 0.13}  & ~~2.69 $\pm$ 0.12       \\ \hline
\end{tabular}
}
\label{tab:results-untargeted-attack}
\end{table*}

\textbf{Untargeted adversarial attacks.}
\Cref{tab:results-untargeted-attack} represents the evaluation of our best models, obtained in 5 training iterations, on attacked test sets using our adversarial algorithms ATA and AGA. 

We achieve consistent attack performance with the best attack SR across all datasets and models using the undefended AGA attack, which features the highest average attack success rates of 43.1\% and 56.09\% for the Swin Transformer V2 model on the Caltech-256 and Food-101 datasets, respectively, and 63.97\% for the ViT model on the Tiny-ImageNet-200 dataset. Thus, we demonstrate that computer vision transformers are weaker against undefended adversarial attacks. The ATA algorithm produces weaker defended and undefended attacks than the AGA algorithm. For Caltech-256, the most successful attacks of AGA have an absolute SR gain of +29.14\% (undefended) and +12.55\% (defended). Similarly, on Food-101, the AGA algorithm conveys +38.4\% and +20.54\%, while on Tiny-ImageNet-200, the algorithm gains +38.17\% and +19.58\%, which is slightly less than the average difference in SR on Food-101. 

In contrast, computer vision transformers achieve a better defense against adversarial data than regular CNN architectures when trained with adversarial data from the same distribution. However, they are significantly affected in the undefended scenario. 

For the Caltech-256 dataset, we achieve the best attack success reduction on the Swin Transformer V2, with a -26.43\% absolute SR drop for AGA and -6.88\% for ATA. The Food-101 dataset features improved defense for Swin Transformer V2 and ViT on the AGA algorithm, with a 34.03\% SR drop and 34.18\%, respectively. The best defense against the ATA algorithm is achieved with the ResNet-18 model, resulting in a 5.89\% decrease in the attack SR. Finally, for the Tiny-ImageNet-200 dataset, the most significant drop is achieved with ViT for AGA, resulting in a substantial absolute drop of -49.6\%, representing the most outstanding defense against our adversarial algorithms in our experiments. For the ATA algorithm, the most significant reduction is observed in the DenseNet-121 model, with a decrease of -12.17\%.

\begin{table}[!htb]

\caption{Targeted attack results. We evaluate our algorithms, Affine Transform Attack (ATA) and Affine Genetic Attack (AGA), for the Caltech-256 dataset, and every model architecture is trained with the regular set and adversarial augmentations. We reflect the results of the attacks through the targeted and untargeted attack success rate (SR), where $\uparrow$ represents a better attack for a higher SR. A negative SR signifies an improvement in model accuracy. The results are presented as the average values from five testing iterations. We depict the experiments for each targeted class (0-9) chosen as the first ten sorted Caltech-256 classes.}
\centering

\resizebox{1.005\columnwidth}{!}{

\begin{tabular}{|c|l|cccc|llll}
\hline
\multirow{3}{*}{\textbf{Model}}   & \multicolumn{1}{c|}{\multirow{3}{*}{\textbf{Target}}} & \multicolumn{4}{c|}{\textbf{Untargeted   SR (\%) $\uparrow$}}                                                                                                                                            & \multicolumn{4}{c|}{\textbf{Targeted   SR (\%) $\uparrow$}}                                                                                                                          \\ \cline{3-10} 
                                  & \multicolumn{1}{c|}{}                                 & \multicolumn{1}{c|}{\multirow{2}{*}{\textbf{AGA}}}    & \multicolumn{1}{c|}{\textbf{AGA}}                     & \multicolumn{1}{c|}{\multirow{2}{*}{\textbf{ATA}}}    & \textbf{ATA}                     & \multicolumn{1}{c|}{\multirow{2}{*}{\textbf{AGA}}} & \multicolumn{1}{c|}{\textbf{AGA}}    & \multicolumn{1}{c|}{\multirow{2}{*}{\textbf{ATA}}} & \multicolumn{1}{c|}{\textbf{ATA}}    \\
                                  & \multicolumn{1}{c|}{}                                 & \multicolumn{1}{c|}{}                                 & \multicolumn{1}{c|}{\textbf{(Aug.)}}                  & \multicolumn{1}{c|}{}                                 & \textbf{(Aug.)}                  & \multicolumn{1}{c|}{}                              & \multicolumn{1}{c|}{\textbf{(Aug.)}} & \multicolumn{1}{c|}{}                              & \multicolumn{1}{c|}{\textbf{(Aug.)}} \\    \hline \hline
\multirow{10}{*}{ResNet-18}        & 0                                                     & \multicolumn{1}{c|}{\multirow{10}{*}{\textbf{52.87}}} & \multicolumn{1}{c|}{\multirow{10}{*}{\textbf{28.90}}} & \multicolumn{1}{c|}{\multirow{10}{*}{\textbf{15.37}}} & \multirow{10}{*}{\textbf{9.59}}  & \multicolumn{1}{l|}{~1.17}                          & \multicolumn{1}{l|}{~1.06}            & \multicolumn{1}{l|}{~0.32}                          & \multicolumn{1}{l|}{~0.20}           \\ \cline{2-2} \cline{7-10} 
                                  & 1                                                     & \multicolumn{1}{c|}{}                                 & \multicolumn{1}{c|}{}                                 & \multicolumn{1}{c|}{}                                 &                                  & \multicolumn{1}{l|}{-0.12}                         & \multicolumn{1}{l|}{~0.82}            & \multicolumn{1}{l|}{~0.28}                          & \multicolumn{1}{l|}{-0.44}          \\ \cline{2-2} \cline{7-10} 
                                  & 2                                                     & \multicolumn{1}{c|}{}                                 & \multicolumn{1}{c|}{}                                 & \multicolumn{1}{c|}{}                                 &                                  & \multicolumn{1}{l|}{-0.52}                         & \multicolumn{1}{l|}{~0.00}            & \multicolumn{1}{l|}{~0.44}                          & \multicolumn{1}{l|}{-0.48}          \\ \cline{2-2} \cline{7-10} 
                                  & 3                                                     & \multicolumn{1}{c|}{}                                 & \multicolumn{1}{c|}{}                                 & \multicolumn{1}{c|}{}                                 &                                  & \multicolumn{1}{l|}{-0.56}                         & \multicolumn{1}{l|}{~0.90}            & \multicolumn{1}{l|}{~0.52}                          & \multicolumn{1}{l|}{-0.08}          \\ \cline{2-2} \cline{7-10} 
                                  & 4                                                     & \multicolumn{1}{c|}{}                                 & \multicolumn{1}{c|}{}                                 & \multicolumn{1}{c|}{}                                 &                                  & \multicolumn{1}{l|}{-0.24}                         & \multicolumn{1}{l|}{-0.45}           & \multicolumn{1}{l|}{~1.16}                          & \multicolumn{1}{l|}{-0.60}          \\ \cline{2-2} \cline{7-10} 
                                  & 5                                                     & \multicolumn{1}{c|}{}                                 & \multicolumn{1}{c|}{}                                 & \multicolumn{1}{c|}{}                                 &                                  & \multicolumn{1}{l|}{~0.52}                          & \multicolumn{1}{l|}{~0.98}            & \multicolumn{1}{l|}{~0.36}                          & \multicolumn{1}{l|}{-0.36}          \\ \cline{2-2} \cline{7-10} 
                                  & 6                                                     & \multicolumn{1}{c|}{}                                 & \multicolumn{1}{c|}{}                                 & \multicolumn{1}{c|}{}                                 &                                  & \multicolumn{1}{l|}{-0.28}                         & \multicolumn{1}{l|}{~0.90}            & \multicolumn{1}{l|}{-0.08}                         & \multicolumn{1}{l|}{-0.36}          \\ \cline{2-2} \cline{7-10} 
                                  & 7                                                     & \multicolumn{1}{c|}{}                                 & \multicolumn{1}{c|}{}                                 & \multicolumn{1}{c|}{}                                 &                                  & \multicolumn{1}{l|}{-0.44}                         & \multicolumn{1}{l|}{~0.37}            & \multicolumn{1}{l|}{~0.28}                         & \multicolumn{1}{l|}{-0.88}          \\ \cline{2-2} \cline{7-10} 
                                  & 8                                                     & \multicolumn{1}{c|}{}                                 & \multicolumn{1}{c|}{}                                 & \multicolumn{1}{c|}{}                                 &                                  & \multicolumn{1}{l|}{-0.44}                         & \multicolumn{1}{l|}{~1.10}            & \multicolumn{1}{l|}{-0.04}                         & \multicolumn{1}{l|}{-0.64}          \\ \cline{2-2} \cline{7-10} 
                                  & 9                                                     & \multicolumn{1}{c|}{}                                 & \multicolumn{1}{c|}{}                                 & \multicolumn{1}{c|}{}                                 &                                  & \multicolumn{1}{l|}{~0.08}                          & \multicolumn{1}{l|}{~0.37}            & \multicolumn{1}{l|}{~0.12}                          & \multicolumn{1}{l|}{-0.56}          \\ \hline \hline
\multirow{10}{*}{DenseNet-121}     & 0                                                     & \multicolumn{1}{c|}{\multirow{10}{*}{\textbf{52.60}}} & \multicolumn{1}{c|}{\multirow{10}{*}{\textbf{30.32}}} & \multicolumn{1}{c|}{\multirow{10}{*}{\textbf{13.95}}} & \multirow{10}{*}{\textbf{10.87}} & \multicolumn{1}{l|}{-0.96}                         & \multicolumn{1}{l|}{~0.20}            & \multicolumn{1}{l|}{~0.04}                          & \multicolumn{1}{l|}{-0.12}          \\ \cline{2-2} \cline{7-10} 
                                  & 1                                                     & \multicolumn{1}{c|}{}                                 & \multicolumn{1}{c|}{}                                 & \multicolumn{1}{c|}{}                                 &                                  & \multicolumn{1}{l|}{-0.39}                         & \multicolumn{1}{l|}{~0.86}            & \multicolumn{1}{l|}{~0.65}                          & \multicolumn{1}{l|}{-0.27}          \\ \cline{2-2} \cline{7-10} 
                                  & 2                                                     & \multicolumn{1}{c|}{}                                 & \multicolumn{1}{c|}{}                                 & \multicolumn{1}{c|}{}                                 &                                  & \multicolumn{1}{l|}{-0.69}                         & \multicolumn{1}{l|}{-0.12}           & \multicolumn{1}{l|}{~0.80}                          & \multicolumn{1}{l|}{~0.50}           \\ \cline{2-2} \cline{7-10} 
                                  & 3                                                     & \multicolumn{1}{c|}{}                                 & \multicolumn{1}{c|}{}                                 & \multicolumn{1}{c|}{}                                 &                                  & \multicolumn{1}{l|}{~0.19}                          & \multicolumn{1}{l|}{-0.04}           & \multicolumn{1}{l|}{~0.65}                          & \multicolumn{1}{l|}{-0.39}          \\ \cline{2-2} \cline{7-10} 
                                  & 4                                                     & \multicolumn{1}{c|}{}                                 & \multicolumn{1}{c|}{}                                 & \multicolumn{1}{c|}{}                                 &                                  & \multicolumn{1}{l|}{-0.46}                         & \multicolumn{1}{l|}{-0.55}           & \multicolumn{1}{l|}{-0.08}                         & \multicolumn{1}{l|}{-0.39}          \\ \cline{2-2} \cline{7-10} 
                                  & 5                                                     & \multicolumn{1}{c|}{}                                 & \multicolumn{1}{c|}{}                                 & \multicolumn{1}{c|}{}                                 &                                  & \multicolumn{1}{l|}{-0.35}                         & \multicolumn{1}{l|}{~0.70}            & \multicolumn{1}{l|}{~0.65}                          & \multicolumn{1}{l|}{-0.08}          \\ \cline{2-2} \cline{7-10} 
                                  & 6                                                     & \multicolumn{1}{c|}{}                                 & \multicolumn{1}{c|}{}                                 & \multicolumn{1}{c|}{}                                 &                                  & \multicolumn{1}{l|}{-0.12}                         & \multicolumn{1}{l|}{-1.13}           & \multicolumn{1}{l|}{~1.22}                          & \multicolumn{1}{l|}{-0.35}          \\ \cline{2-2} \cline{7-10} 
                                  & 7                                                     & \multicolumn{1}{c|}{}                                 & \multicolumn{1}{c|}{}                                 & \multicolumn{1}{c|}{}                                 &                                  & \multicolumn{1}{l|}{-0.50}                         & \multicolumn{1}{l|}{-0.27}           & \multicolumn{1}{l|}{~0.61}                          & \multicolumn{1}{l|}{-0.70}          \\ \cline{2-2} \cline{7-10} 
                                  & 8                                                     & \multicolumn{1}{c|}{}                                 & \multicolumn{1}{c|}{}                                 & \multicolumn{1}{c|}{}                                 &                                  & \multicolumn{1}{l|}{-0.27}                         & \multicolumn{1}{l|}{-1.02}           & \multicolumn{1}{l|}{~0.31}                          & \multicolumn{1}{l|}{~0.16}           \\ \cline{2-2} \cline{7-10} 
                                  & 9                                                     & \multicolumn{1}{c|}{}                                 & \multicolumn{1}{c|}{}                                 & \multicolumn{1}{c|}{}                                 &                                  & \multicolumn{1}{l|}{~0.08}                          & \multicolumn{1}{l|}{~0.08}            & \multicolumn{1}{l|}{~0.69}                          & \multicolumn{1}{l|}{-0.62}          \\ \hline \hline
\multirow{10}{*}{Swin Transformer V2} & 0                                                     & \multicolumn{1}{c|}{\multirow{10}{*}{\textbf{63.41}}} & \multicolumn{1}{c|}{\multirow{10}{*}{\textbf{22.27}}} & \multicolumn{1}{c|}{\multirow{10}{*}{\textbf{12.36}}} & \multirow{10}{*}{\textbf{5.58}}  & \multicolumn{1}{l|}{~4.52}                          & \multicolumn{1}{l|}{~2.36}            & \multicolumn{1}{l|}{~3.10}                          & \multicolumn{1}{l|}{~0.78}           \\ \cline{2-2} \cline{7-10} 
                                  & 1                                                     & \multicolumn{1}{c|}{}                                 & \multicolumn{1}{c|}{}                                 & \multicolumn{1}{c|}{}                                 &                                  & \multicolumn{1}{l|}{~5.65}                          & \multicolumn{1}{l|}{~3.93}            & \multicolumn{1}{l|}{~2.41}                          & \multicolumn{1}{l|}{~1.17}           \\ \cline{2-2} \cline{7-10} 
                                  & 2                                                     & \multicolumn{1}{c|}{}                                 & \multicolumn{1}{c|}{}                                 & \multicolumn{1}{c|}{}                                 &                                  & \multicolumn{1}{l|}{~4.19}                          & \multicolumn{1}{l|}{~2.54}            & \multicolumn{1}{l|}{~2.44}                          & \multicolumn{1}{l|}{~1.39}           \\ \cline{2-2} \cline{7-10} 
                                  & 3                                                     & \multicolumn{1}{c|}{}                                 & \multicolumn{1}{c|}{}                                 & \multicolumn{1}{c|}{}                                 &                                  & \multicolumn{1}{l|}{~7.77}                          & \multicolumn{1}{l|}{~2.93}            & \multicolumn{1}{l|}{~3.97}                          & \multicolumn{1}{l|}{~0.93}           \\ \cline{2-2} \cline{7-10} 
                                  & 4                                                     & \multicolumn{1}{c|}{}                                 & \multicolumn{1}{c|}{}                                 & \multicolumn{1}{c|}{}                                 &                                  & \multicolumn{1}{l|}{~4.48}                          & \multicolumn{1}{l|}{~2.50}            & \multicolumn{1}{l|}{~1.68}                          & \multicolumn{1}{l|}{~0.78}           \\ \cline{2-2} \cline{7-10} 
                                  & 5                                                     & \multicolumn{1}{c|}{}                                 & \multicolumn{1}{c|}{}                                 & \multicolumn{1}{c|}{}                                 &                                  & \multicolumn{1}{l|}{~5.47}                          & \multicolumn{1}{l|}{~2.75}            & \multicolumn{1}{l|}{~2.88}                          & \multicolumn{1}{l|}{~0.82}           \\ \cline{2-2} \cline{7-10} 
                                  & 6                                                     & \multicolumn{1}{c|}{}                                 & \multicolumn{1}{c|}{}                                 & \multicolumn{1}{c|}{}                                 &                                  & \multicolumn{1}{l|}{~3.72}                          & \multicolumn{1}{l|}{~3.57}            & \multicolumn{1}{l|}{~3.35}                          & \multicolumn{1}{l|}{~1.03}           \\ \cline{2-2} \cline{7-10} 
                                  & 7                                                     & \multicolumn{1}{c|}{}                                 & \multicolumn{1}{c|}{}                                 & \multicolumn{1}{c|}{}                                 &                                  & \multicolumn{1}{l|}{~4.30}                          & \multicolumn{1}{l|}{~2.43}            & \multicolumn{1}{l|}{~3.06}                          & \multicolumn{1}{l|}{~0.14}           \\ \cline{2-2} \cline{7-10} 
                                  & 8                                                     & \multicolumn{1}{c|}{}                                 & \multicolumn{1}{c|}{}                                 & \multicolumn{1}{c|}{}                                 &                                  & \multicolumn{1}{l|}{~6.67}                          & \multicolumn{1}{l|}{~2.64}            & \multicolumn{1}{l|}{~3.10}                          & \multicolumn{1}{l|}{-0.04}          \\ \cline{2-2} \cline{7-10} 
                                  & 9                                                     & \multicolumn{1}{c|}{}                                 & \multicolumn{1}{c|}{}                                 & \multicolumn{1}{c|}{}                                 &                                  & \multicolumn{1}{l|}{~5.61}                          & \multicolumn{1}{l|}{~2.21}            & \multicolumn{1}{l|}{~2.77}                          & \multicolumn{1}{l|}{~0.64}           \\ \hline \hline
\multirow{10}{*}{ViT}             & 0                                                     & \multicolumn{1}{c|}{\multirow{10}{*}{\textbf{43.82}}} & \multicolumn{1}{c|}{\multirow{10}{*}{\textbf{15.28}}} & \multicolumn{1}{c|}{\multirow{10}{*}{\textbf{4.94}}}  & \multirow{10}{*}{\textbf{3.35}}  & \multicolumn{1}{l|}{~1.20}                          & \multicolumn{1}{l|}{~0.14}            & \multicolumn{1}{l|}{~0.24}                          & \multicolumn{1}{l|}{~0.58}           \\ \cline{2-2} \cline{7-10} 
                                  & 1                                                     & \multicolumn{1}{c|}{}                                 & \multicolumn{1}{c|}{}                                 & \multicolumn{1}{c|}{}                                 &                                  & \multicolumn{1}{l|}{~0.58}                          & \multicolumn{1}{l|}{-0.21}           & \multicolumn{1}{l|}{~0.79}                          & \multicolumn{1}{l|}{~0.62}           \\ \cline{2-2} \cline{7-10} 
                                  & 2                                                     & \multicolumn{1}{c|}{}                                 & \multicolumn{1}{c|}{}                                 & \multicolumn{1}{c|}{}                                 &                                  & \multicolumn{1}{l|}{~0.55}                          & \multicolumn{1}{l|}{-0.03}           & \multicolumn{1}{l|}{~0.79}                          & \multicolumn{1}{l|}{~0.55}           \\ \cline{2-2} \cline{7-10} 
                                  & 3                                                     & \multicolumn{1}{c|}{}                                 & \multicolumn{1}{c|}{}                                 & \multicolumn{1}{c|}{}                                 &                                  & \multicolumn{1}{l|}{~1.27}                          & \multicolumn{1}{l|}{-0.10}           & \multicolumn{1}{l|}{~0.93}                          & \multicolumn{1}{l|}{~0.10}           \\ \cline{2-2} \cline{7-10} 
                                  & 4                                                     & \multicolumn{1}{c|}{}                                 & \multicolumn{1}{c|}{}                                 & \multicolumn{1}{c|}{}                                 &                                  & \multicolumn{1}{l|}{~0.58}                          & \multicolumn{1}{l|}{~0.62}            & \multicolumn{1}{l|}{~0.55}                          & \multicolumn{1}{l|}{~0.48}           \\ \cline{2-2} \cline{7-10} 
                                  & 5                                                     & \multicolumn{1}{c|}{}                                 & \multicolumn{1}{c|}{}                                 & \multicolumn{1}{c|}{}                                 &                                  & \multicolumn{1}{l|}{~1.41}                          & \multicolumn{1}{l|}{~0.10}            & \multicolumn{1}{l|}{~0.79}                          & \multicolumn{1}{l|}{~0.48}           \\ \cline{2-2} \cline{7-10} 
                                  & 6                                                     & \multicolumn{1}{c|}{}                                 & \multicolumn{1}{c|}{}                                 & \multicolumn{1}{c|}{}                                 &                                  & \multicolumn{1}{l|}{~0.75}                          & \multicolumn{1}{l|}{~0.14}            & \multicolumn{1}{l|}{~0.79}                          & \multicolumn{1}{l|}{~0.51}           \\ \cline{2-2} \cline{7-10} 
                                  & 7                                                     & \multicolumn{1}{c|}{}                                 & \multicolumn{1}{c|}{}                                 & \multicolumn{1}{c|}{}                                 &                                  & \multicolumn{1}{l|}{~0.31}                          & \multicolumn{1}{l|}{~0.00}            & \multicolumn{1}{l|}{~0.55}                          & \multicolumn{1}{l|}{~0.58}           \\ \cline{2-2} \cline{7-10} 
                                  & 8                                                     & \multicolumn{1}{c|}{}                                 & \multicolumn{1}{c|}{}                                 & \multicolumn{1}{c|}{}                                 &                                  & \multicolumn{1}{l|}{~0.72}                          & \multicolumn{1}{l|}{-0.10}           & \multicolumn{1}{l|}{~0.96}                          & \multicolumn{1}{l|}{~0.34}           \\ \cline{2-2} \cline{7-10} 
                                  & 9                                                     & \multicolumn{1}{c|}{}                                 & \multicolumn{1}{c|}{}                                 & \multicolumn{1}{c|}{}                                 &                                  & \multicolumn{1}{l|}{~0.48}                          & \multicolumn{1}{l|}{-0.07}           & \multicolumn{1}{l|}{~0.82}                          & \multicolumn{1}{l|}{~0.58}           \\ \hline
\end{tabular}

}
\label{tab:annex-targeted-attack}
\end{table}

\begin{figure*}[!hbt]
\centering
\newcommand{\subfigsize}{0.2475\textwidth}
\newcommand{\tabhsize}{\hspace{-2.75em}}
\newcommand{\tabvsize}{0em}
\newcommand{\firstcolumnadjust}{-2.45}
\newcommand{\firstrowadjust}{0.5}
\tiny
\begin{tabular}{c >{\hspace{-2.2em}} c >{\tabhsize} c >{\tabhsize} c >{\tabhsize} c >{\tabhsize} c}
    & ATA & ATA-Augmented & AGA & AGA-Augmented  \\

    \multirow{-3.5}{*}{\rotatebox[origin=c]{90}{ResNet-18}} & 
    \includegraphics[width=\subfigsize,valign=m]{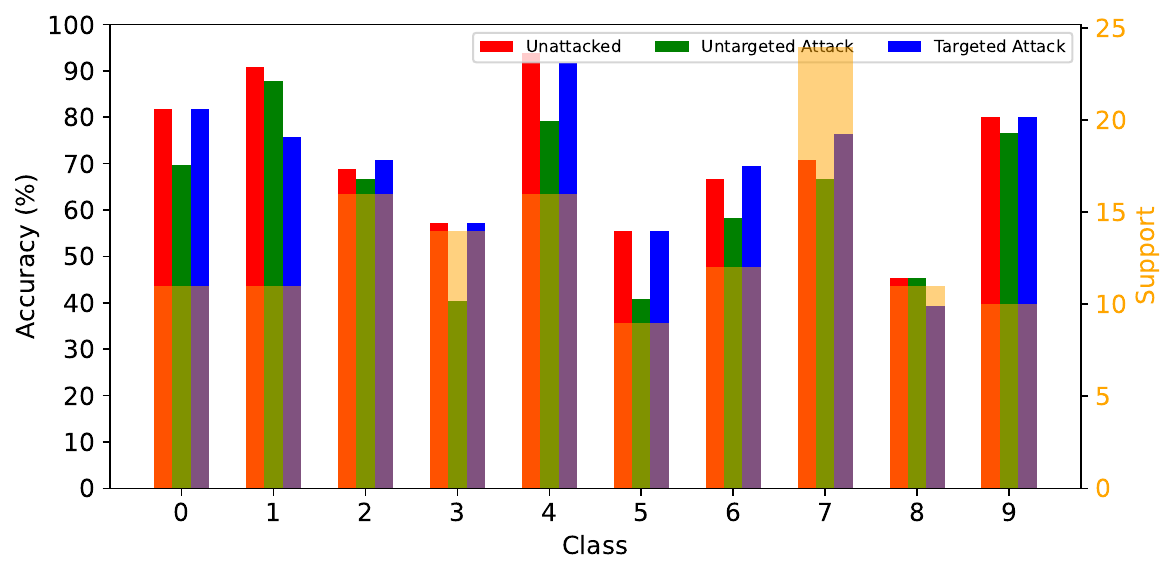} &
    \includegraphics[width=\subfigsize,valign=m]{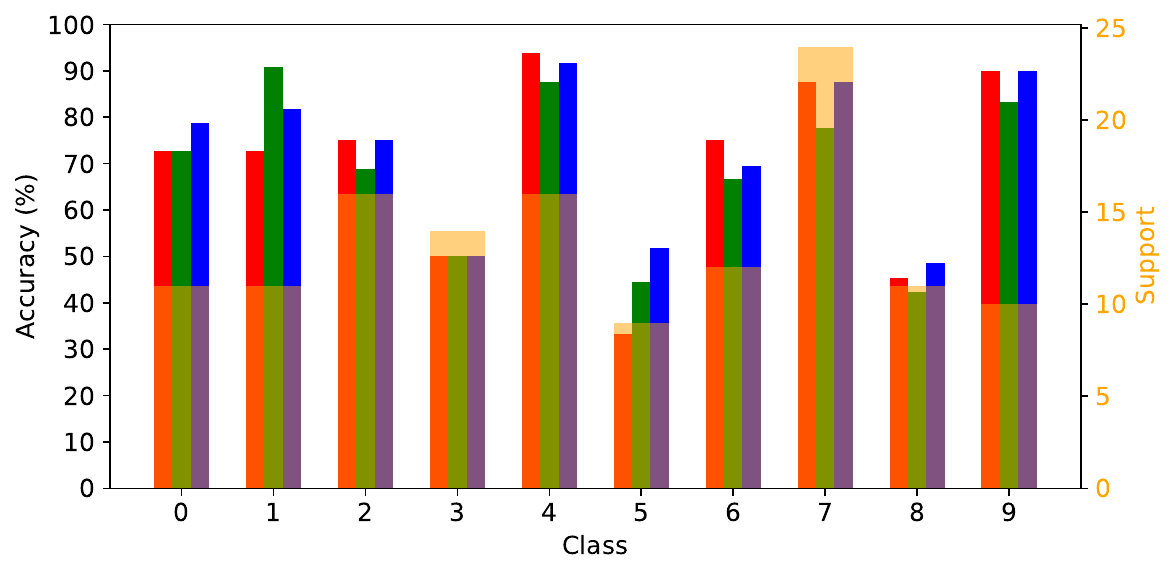} &
    \includegraphics[width=\subfigsize,valign=m]{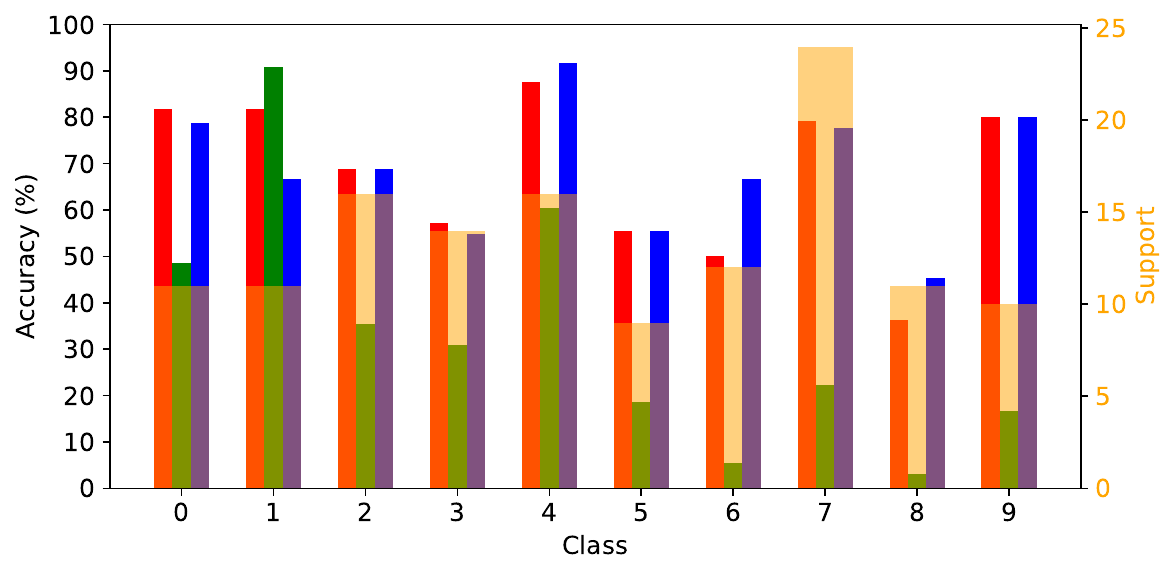} &
    \includegraphics[width=\subfigsize,valign=m]{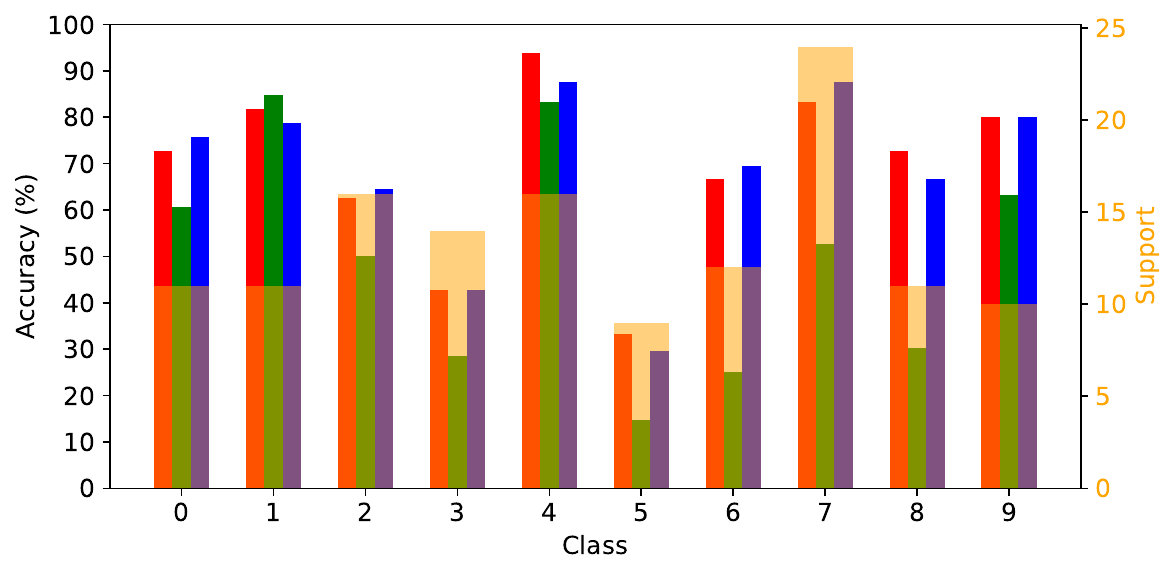} \\ [\tabvsize]

    \multirow{-4.2}{*}{\rotatebox[origin=c]{90}{DenseNet-121}} & 
    \includegraphics[width=\subfigsize,valign=m]{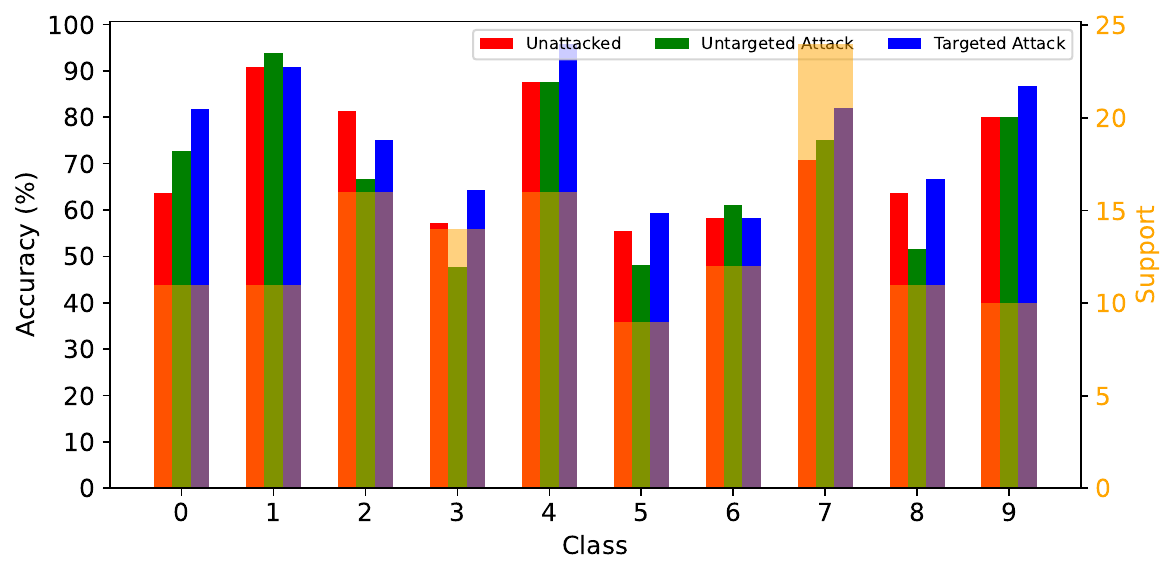} &
    \includegraphics[width=\subfigsize,valign=m]{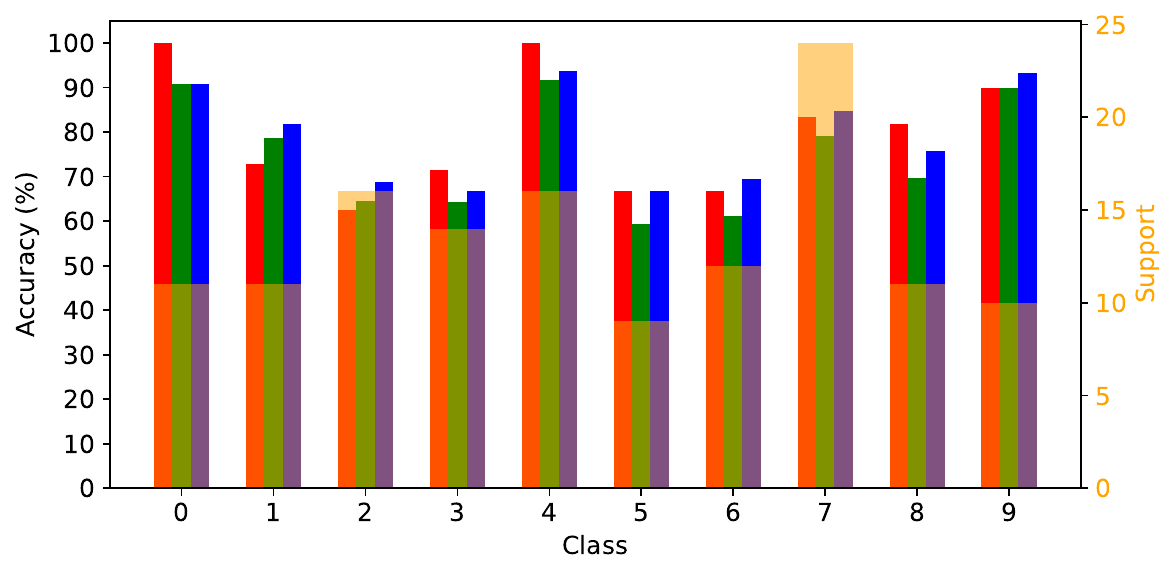} &
    \includegraphics[width=\subfigsize,valign=m]{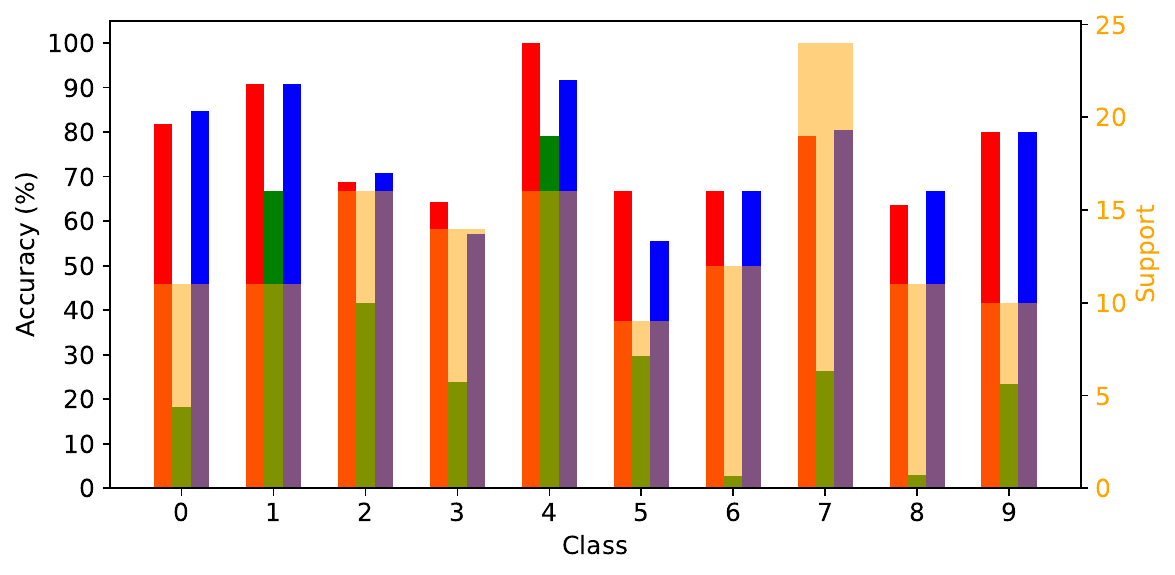} &
    \includegraphics[width=\subfigsize,valign=m]{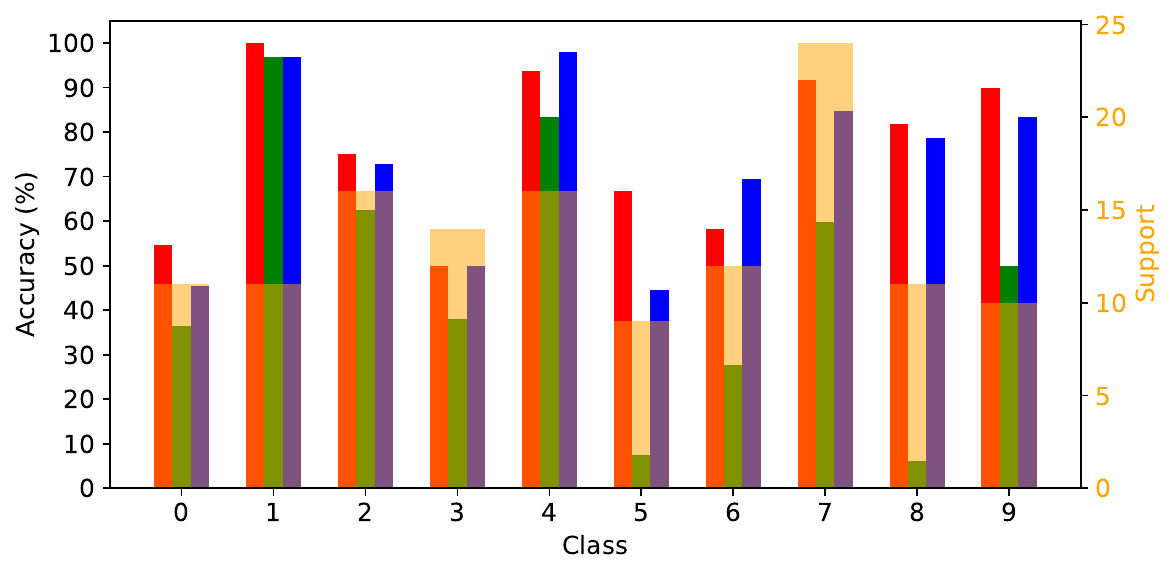} \\ [\tabvsize]

    \multirow{-5.8}{*}{\rotatebox[origin=c]{90}{Swin Transformer V2}} & 
    \includegraphics[width=\subfigsize,valign=m]{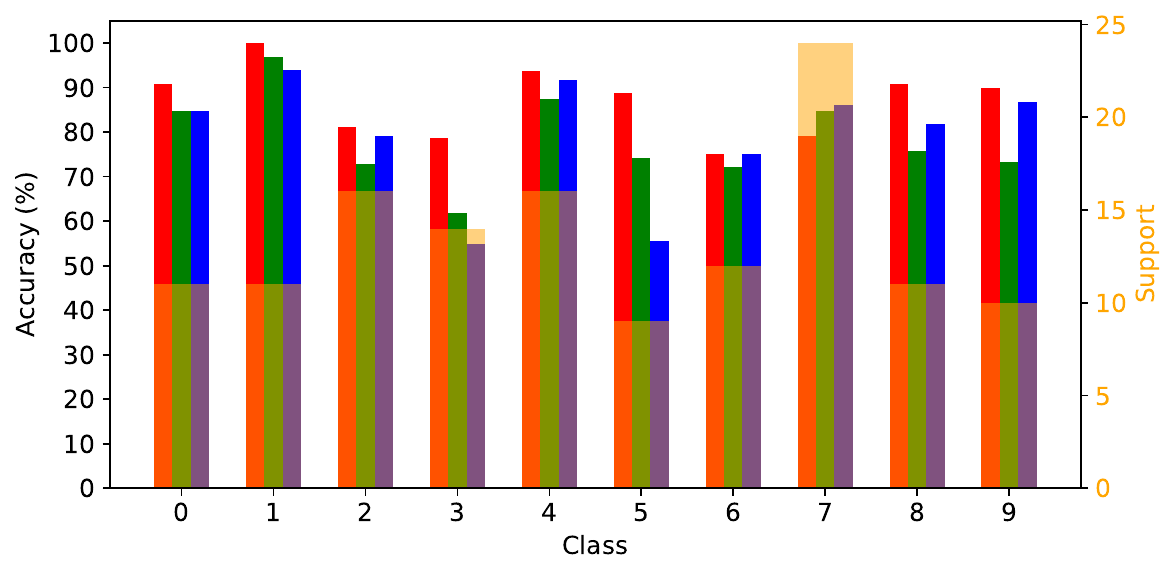} &
    \includegraphics[width=\subfigsize,valign=m]{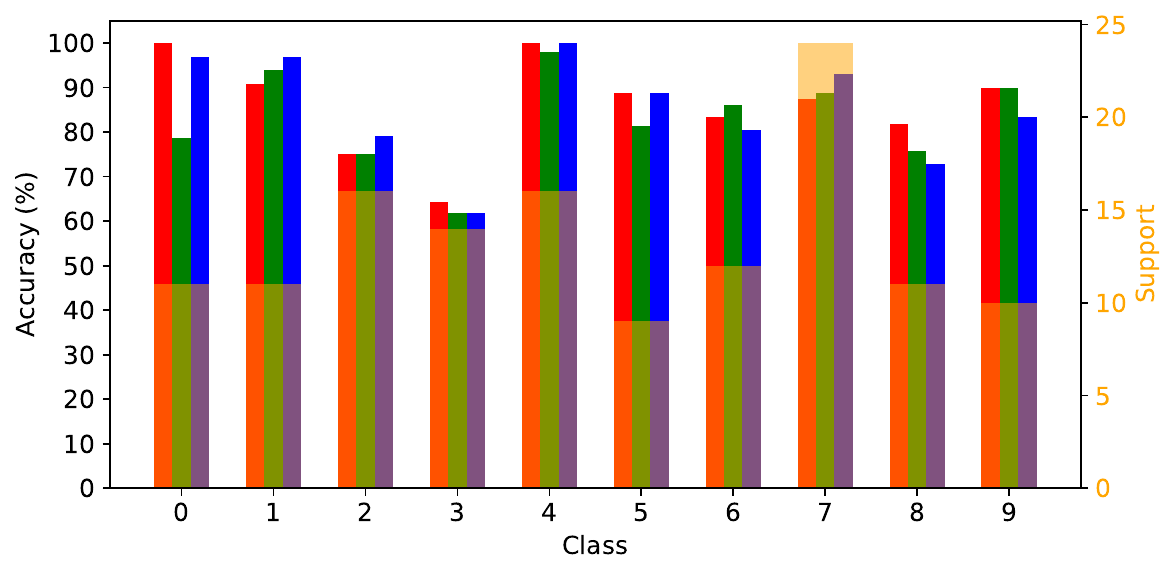} &
    \includegraphics[width=\subfigsize,valign=m]{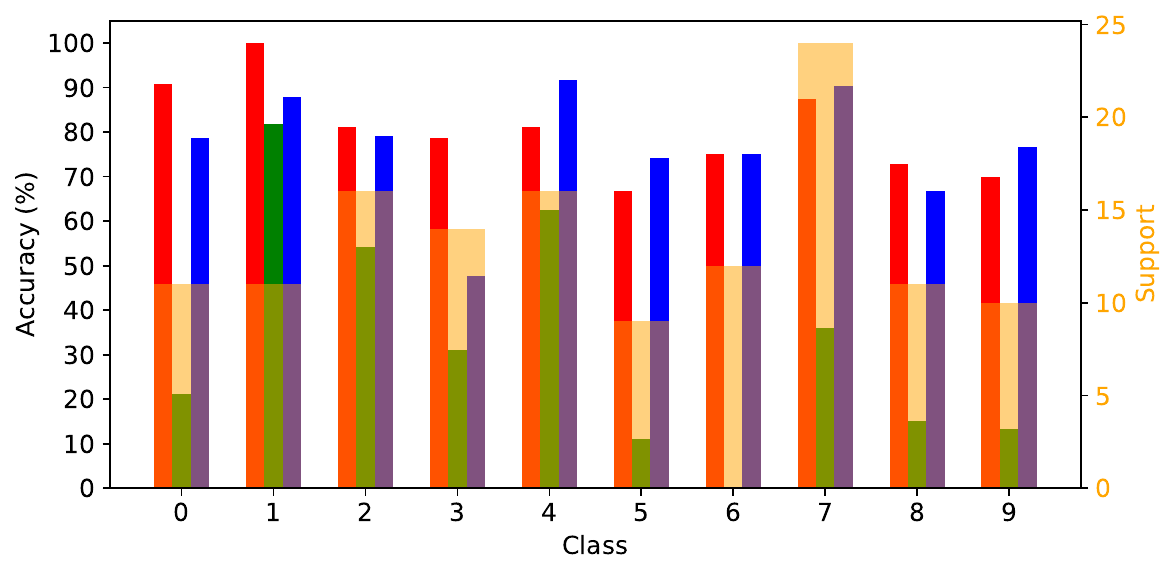} &
    \includegraphics[width=\subfigsize,valign=m]{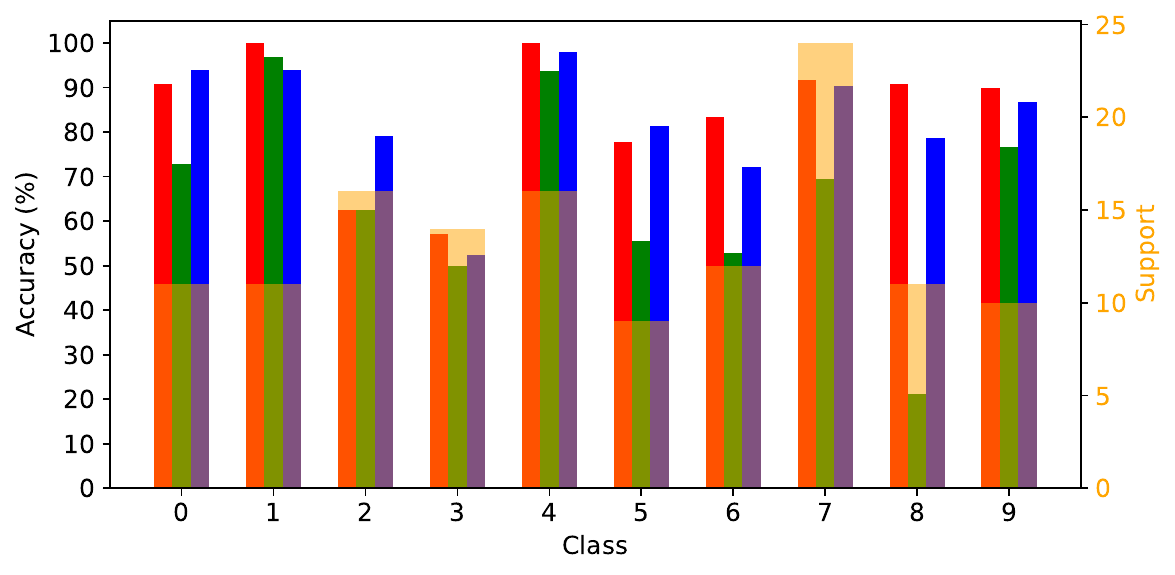} \\ [\tabvsize]

    \multirow{-2.5}{*}{\rotatebox[origin=c]{90}{ViT}} & 
    \includegraphics[width=\subfigsize,valign=m]{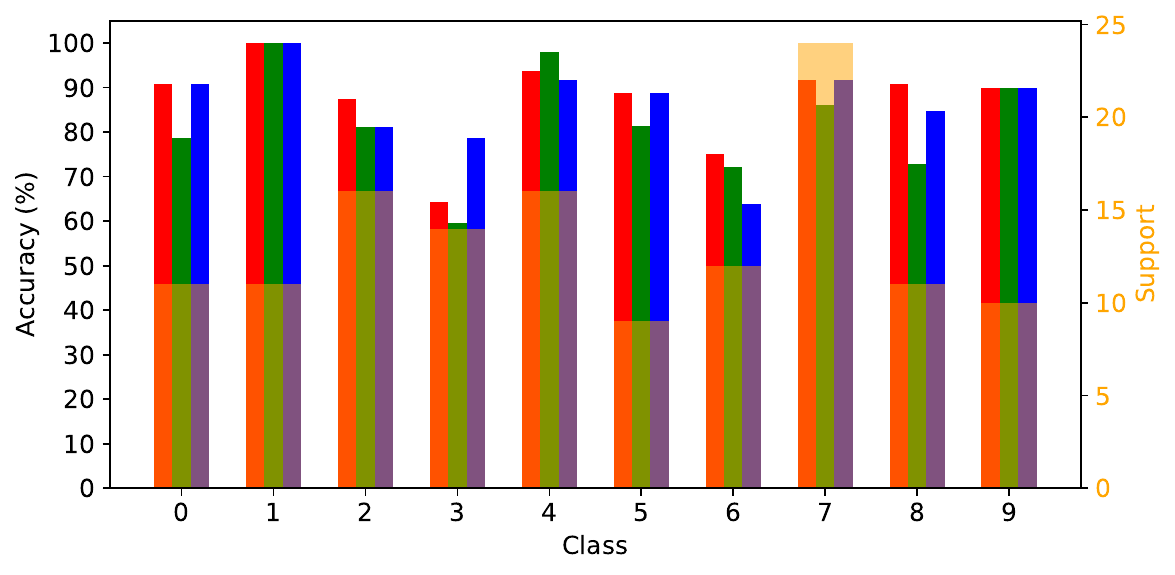} &
    \includegraphics[width=\subfigsize,valign=m]{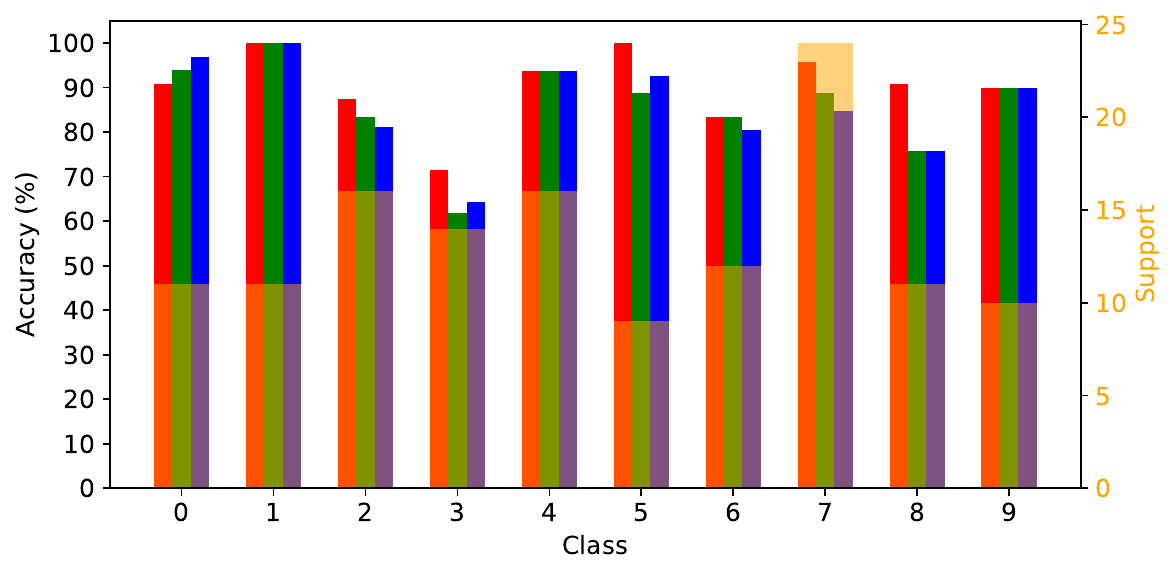} &
    \includegraphics[width=\subfigsize,valign=m]{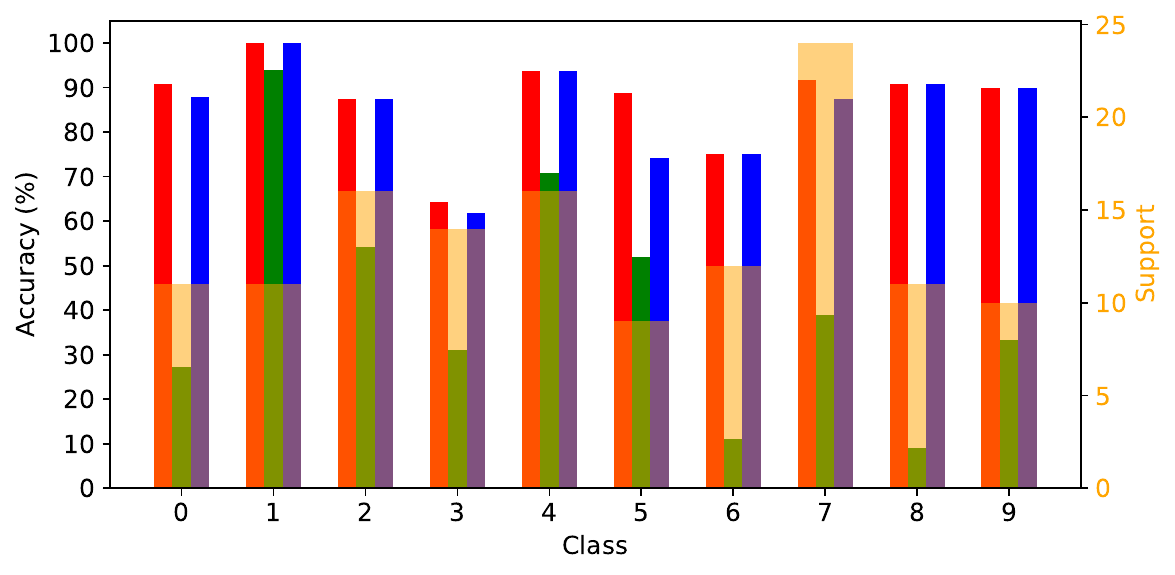} &
    \includegraphics[width=\subfigsize,valign=m]{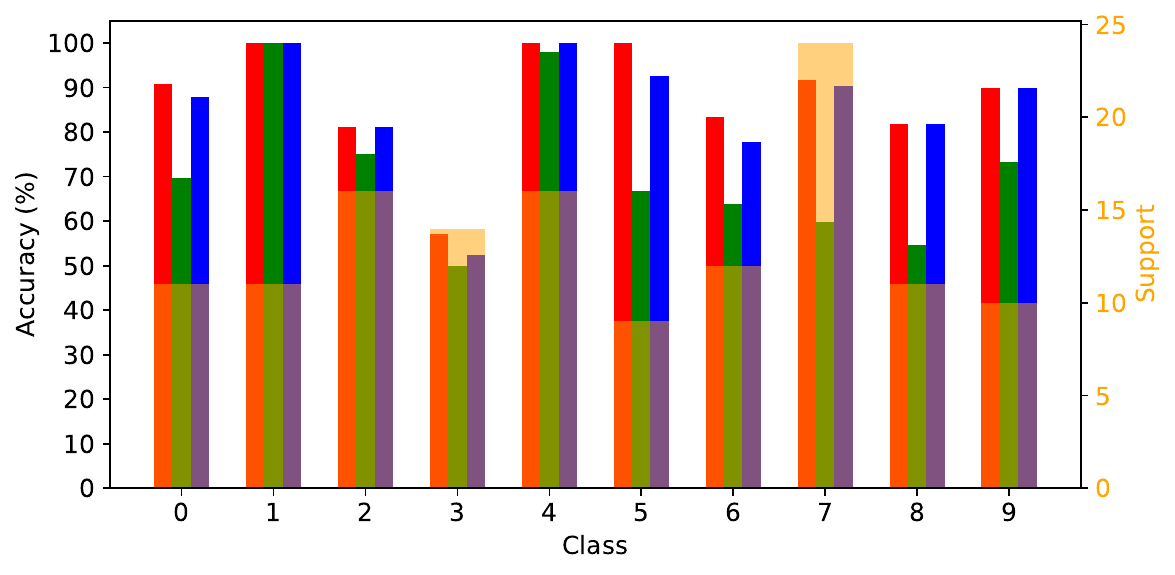} \\ [\tabvsize]

\end{tabular}

\caption{Targeted Attacks for the first ten classes in the Caltech-256 dataset, with ResNet-18, DenseNet-121, ViT, and Swin Transformer V2 architectures. The classes are sorted alphabetically. The red bars represent the accuracy for regular test set evaluation, and the green and blue bars depict the accuracy for the untargeted and targeted attack evaluation, respectively. The orange bars denote the number of images per class.}
\label{fig:results-targeted-attack}
\end{figure*}

\textbf{Targeted adversarial attacks.}
\Cref{tab:annex-targeted-attack} presents the results of targeted adversarial attacks for different classes of the Caltech-256 dataset, specifically the first ten sorted ones, tested on our models. Visibly, the untargeted attacks achieve more model confusion for every model. We merge the untargeted SR values for every target class depicted in \cref{tab:annex-targeted-attack}, as they represent the same experiment for each class.
Intuitively, for these untargeted experiments, we observe a better attack SR for our models in the following order: AGA, AGA-Augmented, ATA, and ATA-Augmented. The ATA's most significant attack SR is achieved for the ResNet-18 architecture, with 15.37\%. The DenseNet-121 features the highest untargeted SR for ATA and AGA augmented models, with 10.87\% and 30.32\%, respectively. Finally, the AGA algorithm achieves the best results against the regularly trained Swin Transformer V2, with a 63.41\% SR.
In contrast, all targeted adversarial attacks achieve a maximum attack SR of 7.77\%. For the ResNet-18 architecture, we determine a maximum SR of 1.17\% on the target class 0 (``001.ak47'') with the AGA model and a minimum SR of -0.88\% on the ATA-Augmented model for class 7 (``008.bathtub''). The DenseNet-121 architecture reaches a maximum of 1.22\% SR on the ATA model for class 6 (``007.bat'') and a minimum of -1.13\% SR on the AGA-Augmented model for the same class. Moreover, the Swin Transformer V2 is the most vulnerable to targeted attacks, with a global maximum of 7.77\% SR on the AGA model for class 3 (``004.baseball-bat'') and a minimum of -0.04\% SR on the ATA-Augmented model for class 8 (``009.bear''). Finally, the ViT architecture shows a slight improvement in targeted attack resistance for class focus in the transformer architectures, with a maximum of 1.41\% SR for the AGA model on class 5 (``006.basketball-hoop'') and a minimum of -0.21\% SR on class 1 (``002.american-flag'') for the AGA-Augmented model. 

Based on our results, we determine that there is no correlation between the SR score and a specific target class. However, considering the experimental results of the transformer architectures, we find that they exhibit a limited improvement in the classification accuracy (negative SR) for specific class targets. Comparing the convolutional results with the transformer ones, we obtain for the latter up to 6.55\% more SR and a -0.92\% lower absolute classification improvement.

\Cref{fig:results-targeted-attack} depicts the resulting average accuracy for the first ten sorted classes in the Caltech-256 dataset. For ResNet-18 models, we attain better targeted classification against unattacked experiments and undefended attacks for 3/10 classes in the ATA, ATA-Augmented, and AGA experiments. For the AGA-Augmented experiment, we obtain an improvement in 4/10 classes. 
The ViT models feature less sensitivity to the influence of the target class. The results show only one improved classification for ATA and ATA-Augmented, and no improvement on the AGA and AGA-Augmented experiments. However, for the latter two experiments, compared to the undefended attack classification, we observe AGA and AGA-Augmented accuracy gains for all classes, except for class 1 of AGA-Augmented, where the accuracy remains the same across all three types of experiments. 
Compared to the results of ResNet-18 and ViT, we achieve better model target influence with DenseNet-121 and Swin Transformer V2. DenseNet-121 shows classification improvements on 7/10 and 5/10 classes for the ATA and ATA-Augmented models, and a slightly reduced improvement on the AGA and AGA-Augmented models, with 4/10 and 2/10 classes, respectively. 
Despite the more considerable attack impact of the AGA algorithm, we gain more significant improvements on the Swin Transformer V2, attaining 4/10 classification improvements for the regular AGA and 3/10 for the AGA-Augmented. In contrast, we achieve only a 1/10 class improvement on the regular (unattacked) model compared to the ATA model, while the ATA-Augmented model had a better influence on 3/10 classes. 
Considering the previous results and model architecture types, we conclude that DenseNet-121 has a greater class influence compared to ResNet-18, with five additional improved classes. Similarly, the Swin Transformer V2 features slightly better targeted attacks, with eight improved classes over the ViT architecture.

\subsection{Algorithm Parameter Variation}

\begin{figure}[!htb]
\centering
\newcommand{\subfigsize}{0.1129\textwidth} 
\newcommand{\tabhsize}{\hspace{-2.5em}}
\newcommand{\tabvsize}{0em}
\newcommand{\firstcolumnadjust}{-3.8}
\newcommand{\firstrowadjust}{1}
\tiny
\begin{tabular}{c >{\hspace{-3.5em}} c >{\hspace{-3.5em}} c >{\tabhsize} c >{\tabhsize} c >{\tabhsize} c >{\tabhsize} c}
    & & \# of Iterations ($n_i$) & Cross. Probability ($p_c$) & Mut. Probability ($p_m$) & Noise Intensity ($\epsilon$) \\

    \multirow{-3.2}{*}{\rotatebox[origin=c]{90}{Caltech-256}} & 
    \multirow{-3.8}{*}{\rotatebox[origin=c]{90}{Attack SR (\%)}} & 
    \includegraphics[width=\subfigsize,valign=m]{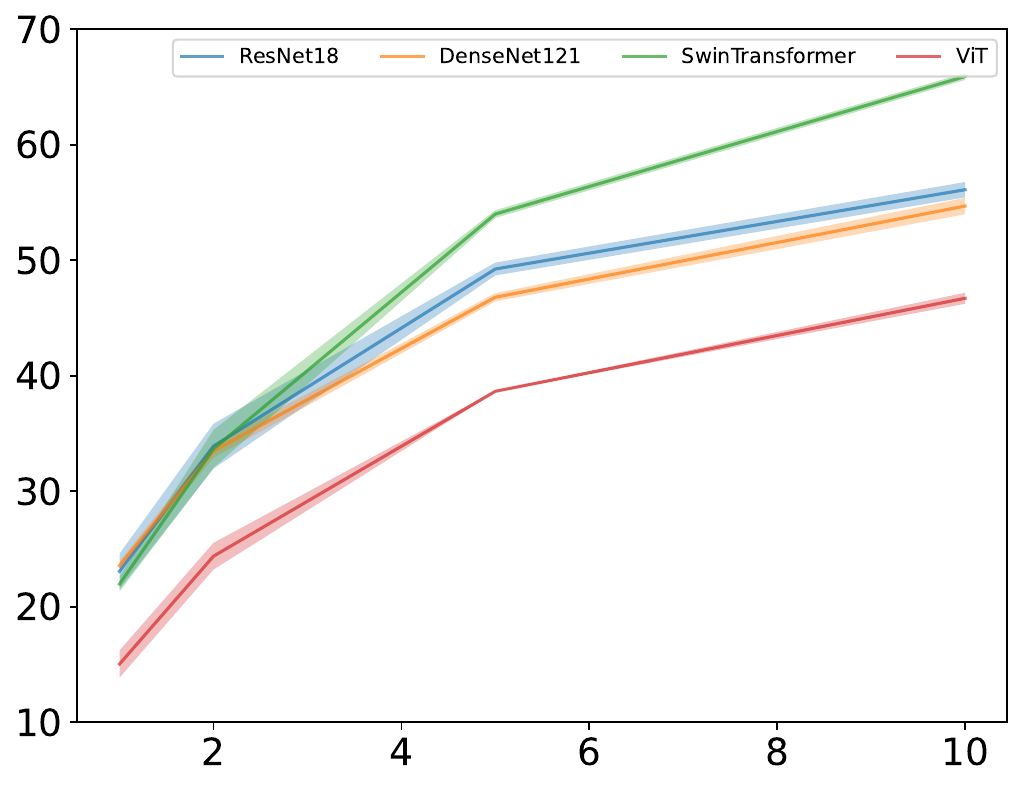} &
    \includegraphics[width=\subfigsize,valign=m]{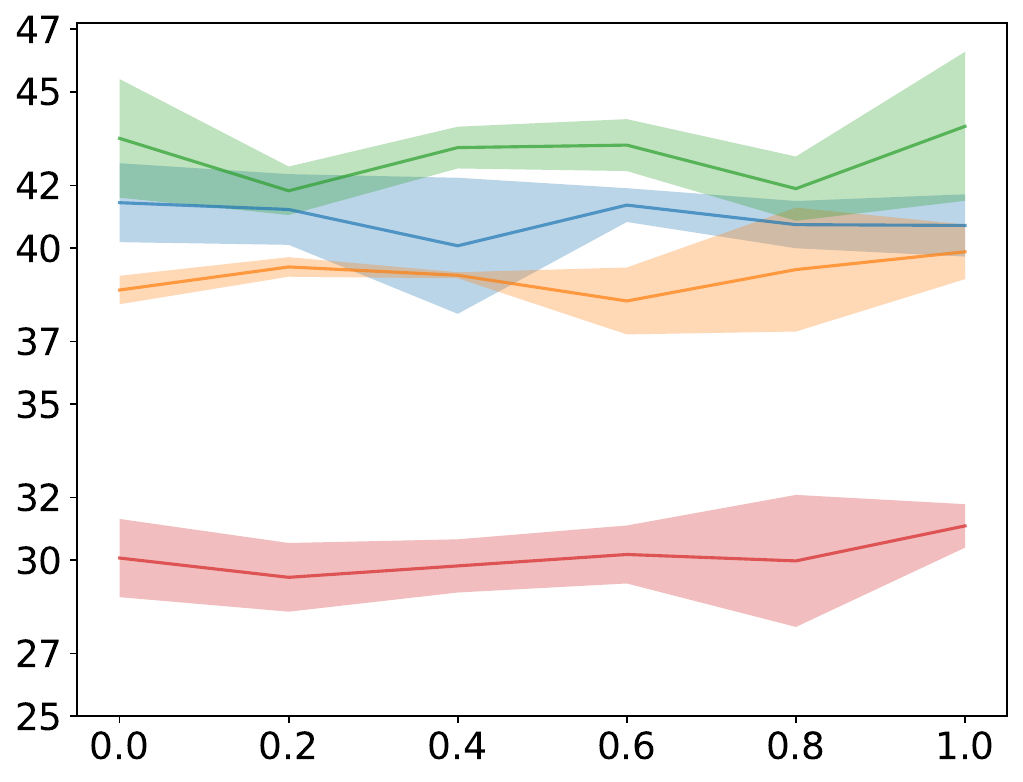} &
    \includegraphics[width=\subfigsize,valign=m]{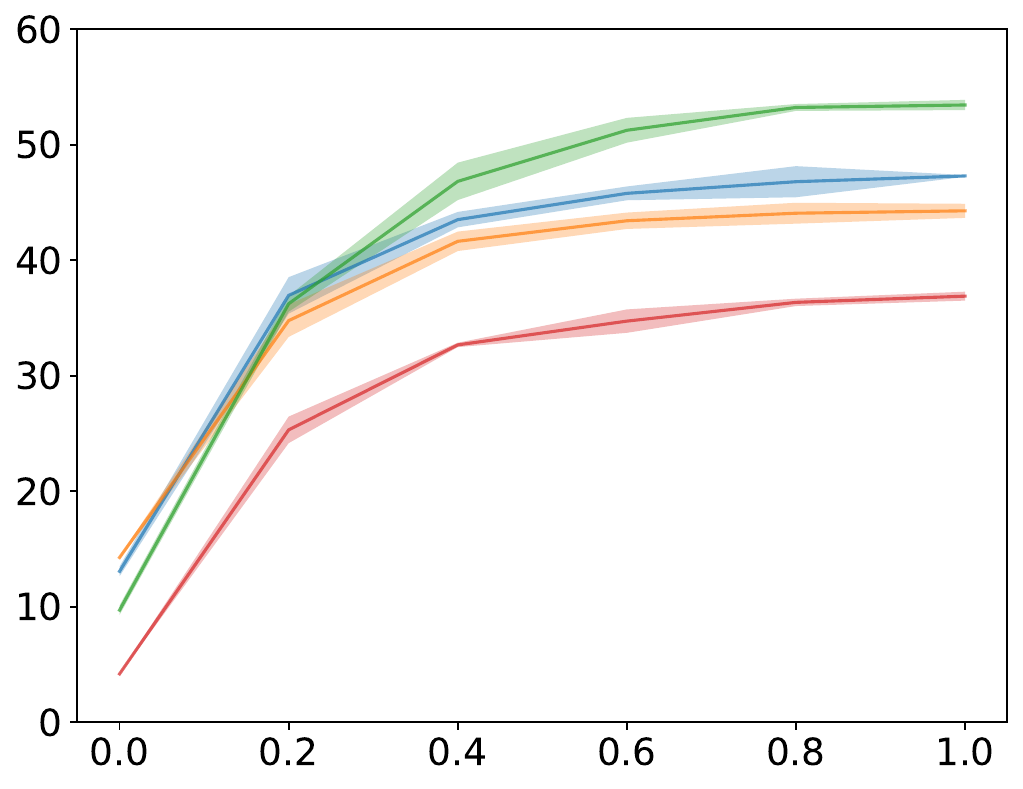} &
    \includegraphics[width=\subfigsize,valign=m]{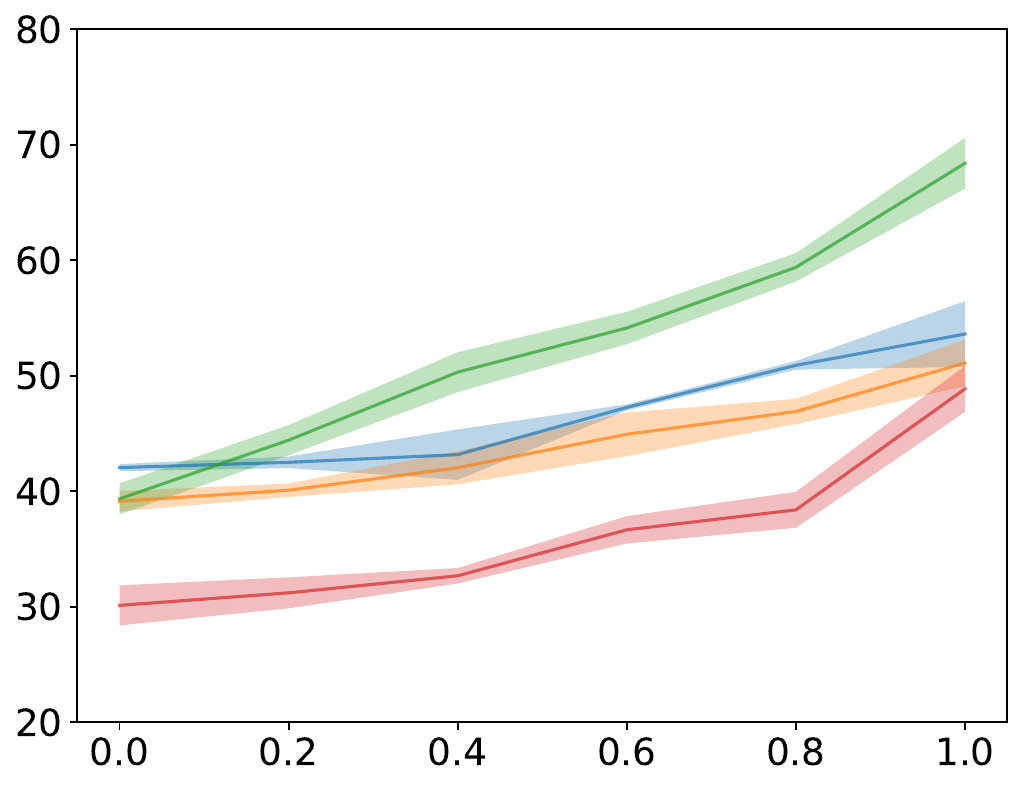} \\ [\tabvsize]

    \multirow{-2.7}{*}{\rotatebox[origin=c]{90}{Food-101}} &
    \multirow{-3.6}{*}{\rotatebox[origin=c]{90}{Attack SR (\%)}} & 
    \includegraphics[width=\subfigsize,valign=m]{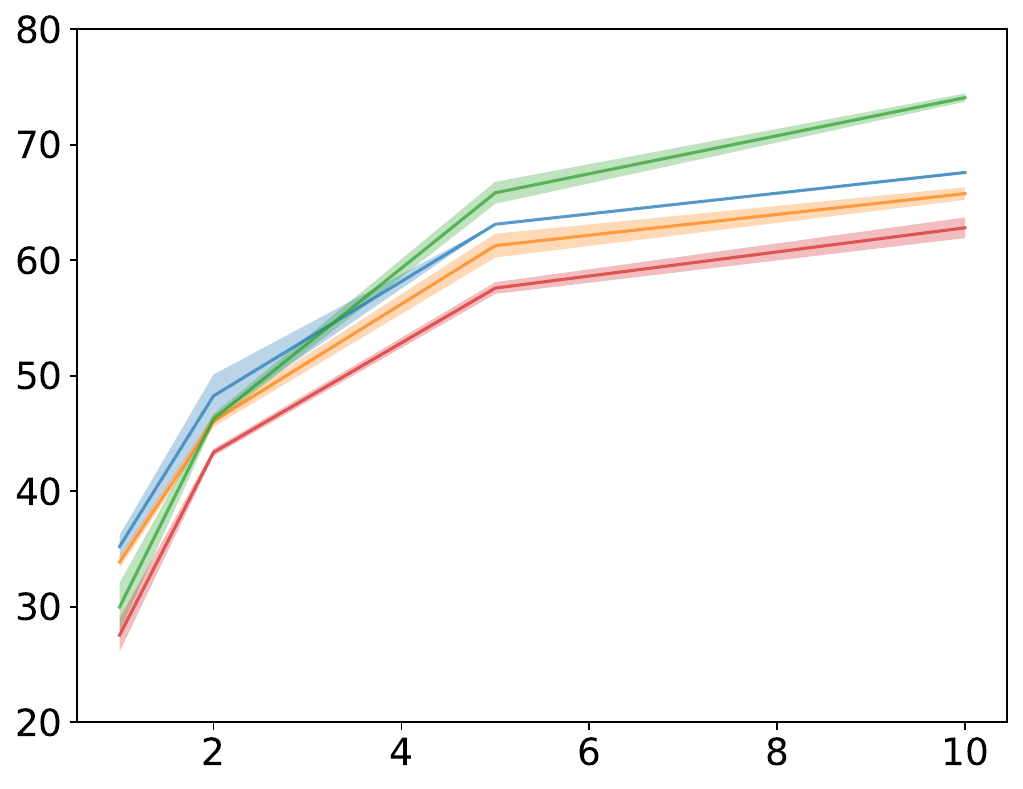} &
    \includegraphics[width=\subfigsize,valign=m]{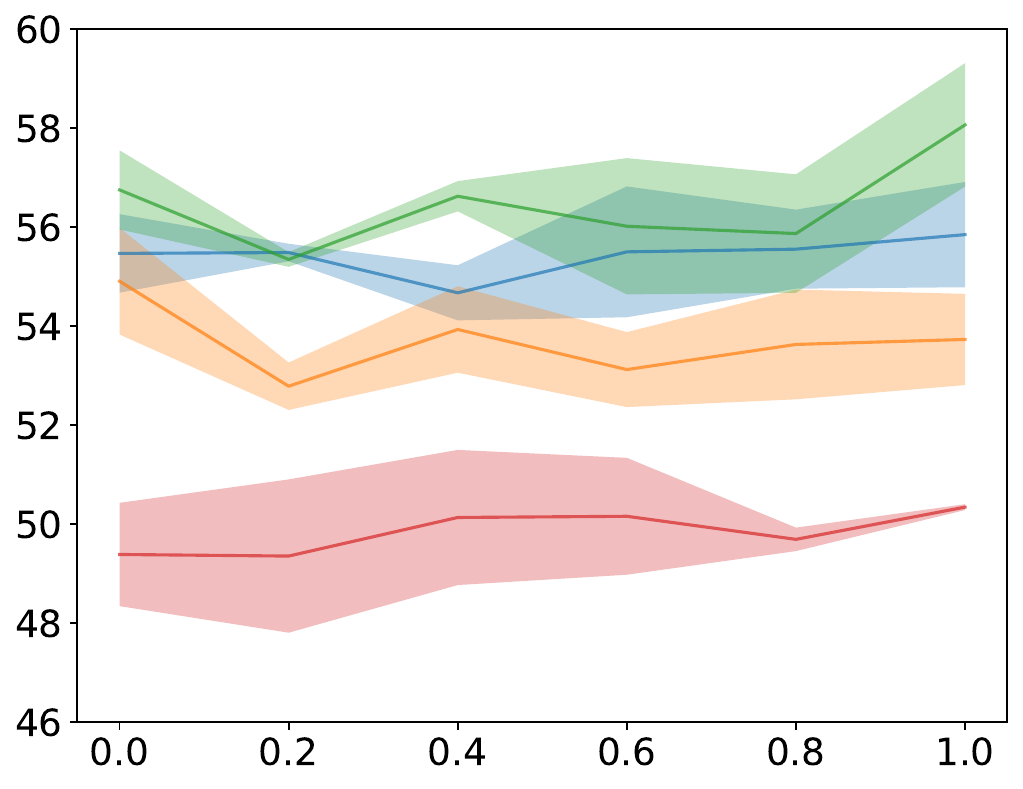} &
    \includegraphics[width=\subfigsize,valign=m]{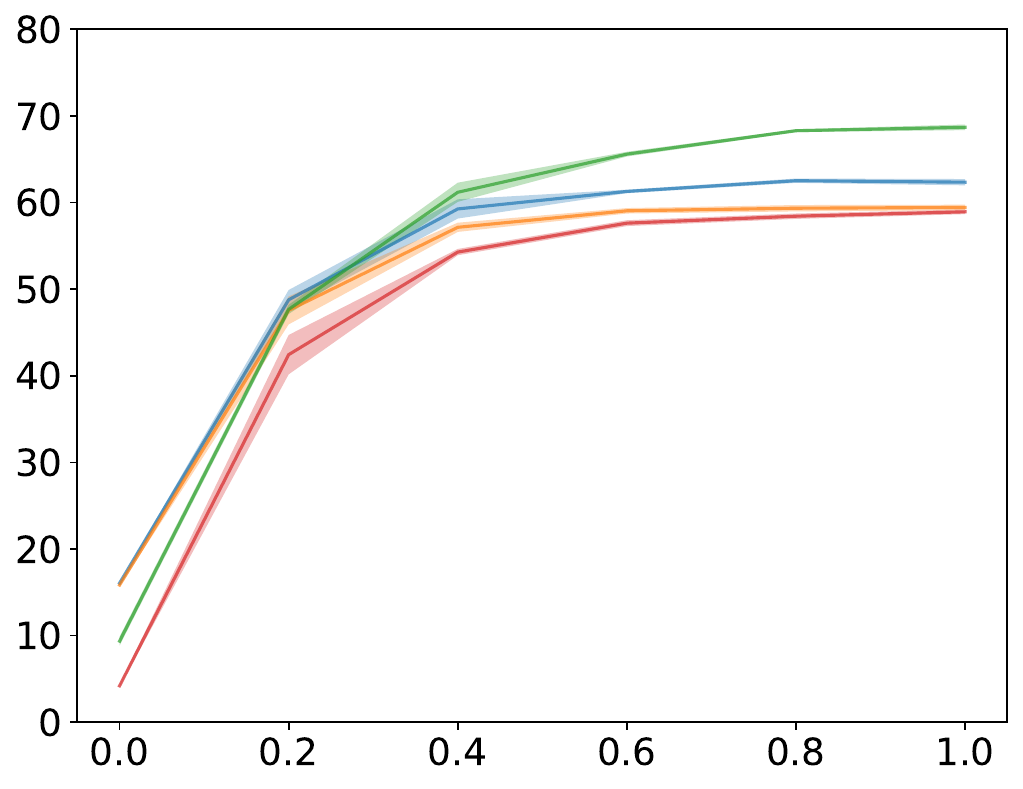} &
    \includegraphics[width=\subfigsize,valign=m]{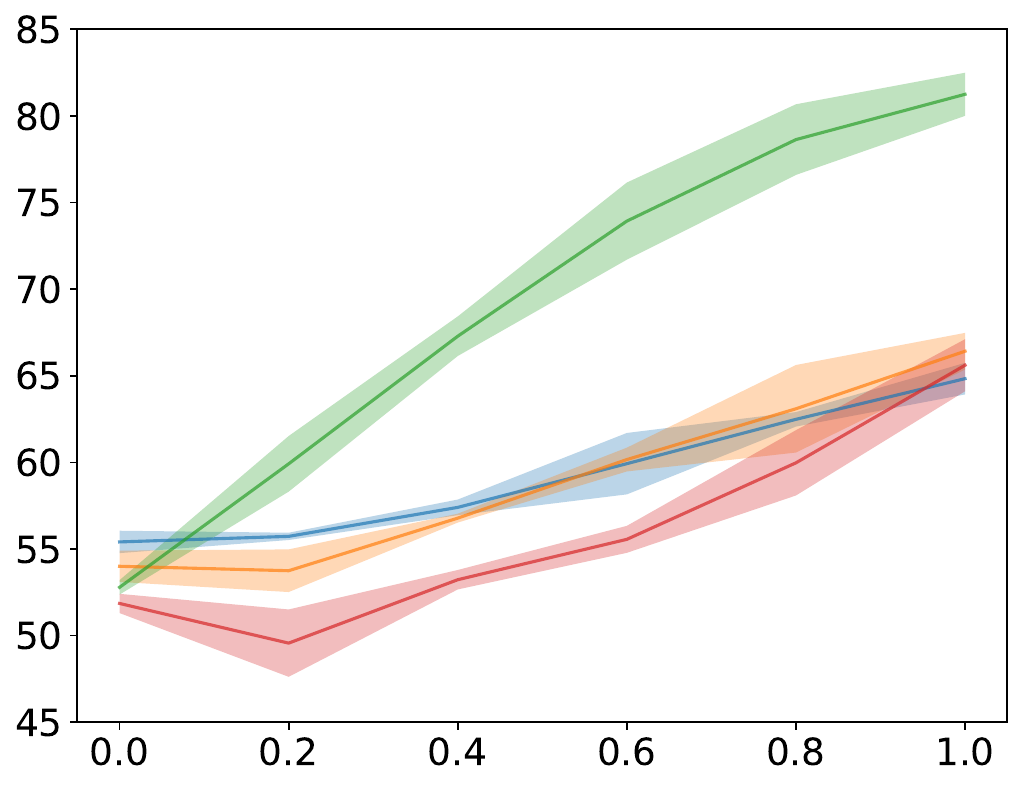} \\ [\tabvsize]

    \multirow{-4.9}{*}{\rotatebox[origin=c]{90}{Tiny-ImageNet-200}} & 
    \multirow{-3.6}{*}{\rotatebox[origin=c]{90}{Attack SR (\%)}} & 
    \includegraphics[width=\subfigsize,valign=m]{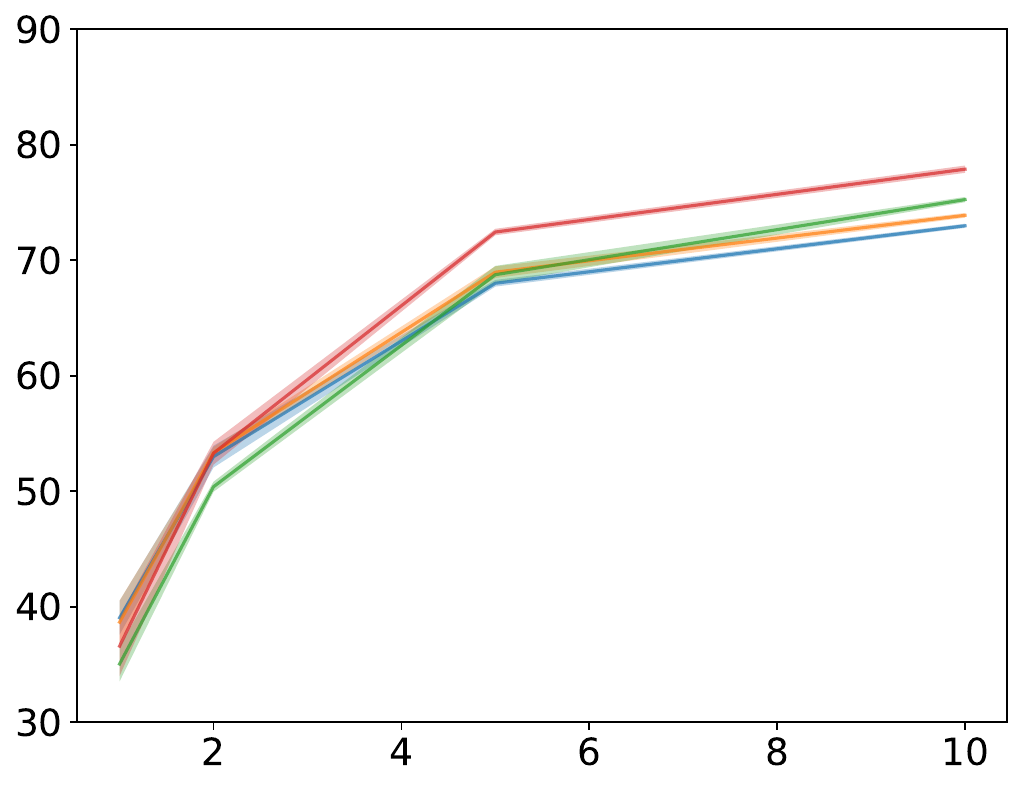} &
    \includegraphics[width=\subfigsize,valign=m]{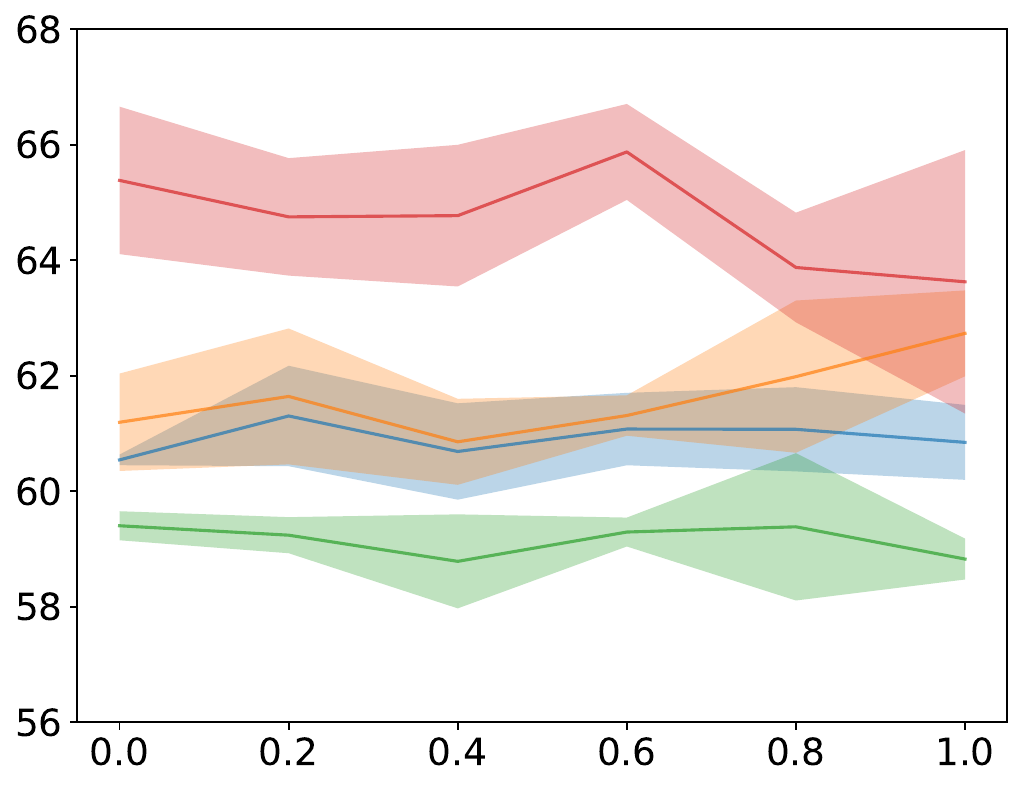} &
    \includegraphics[width=\subfigsize,valign=m]{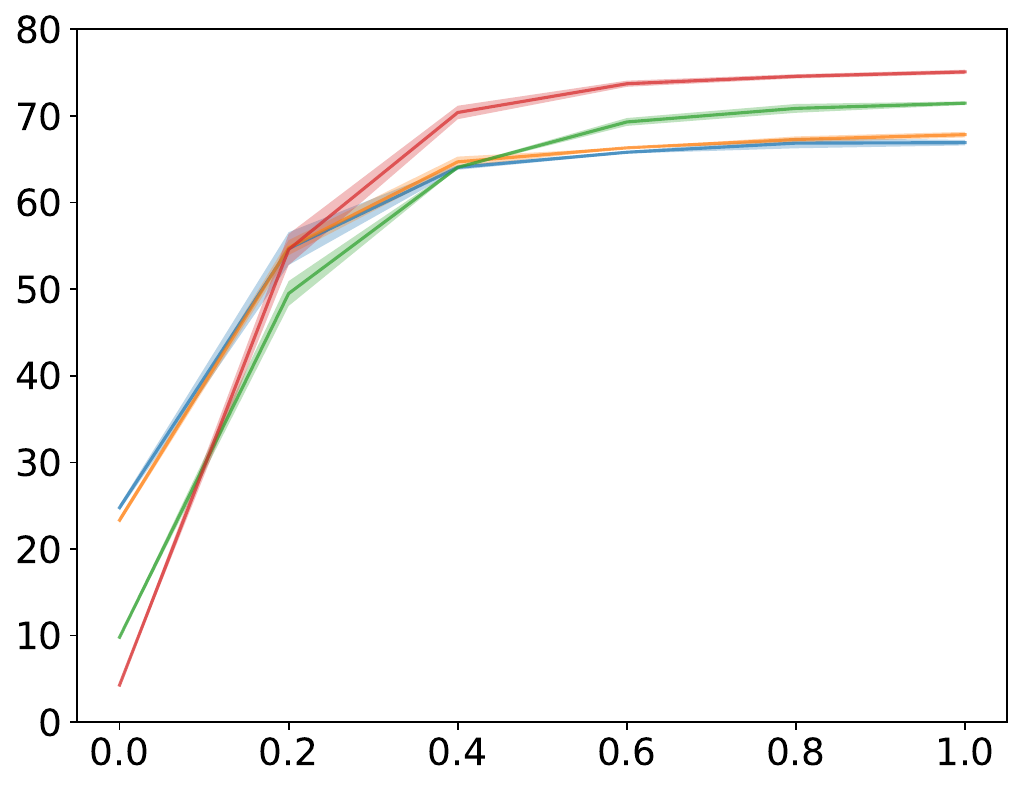} &
    \includegraphics[width=\subfigsize,valign=m]{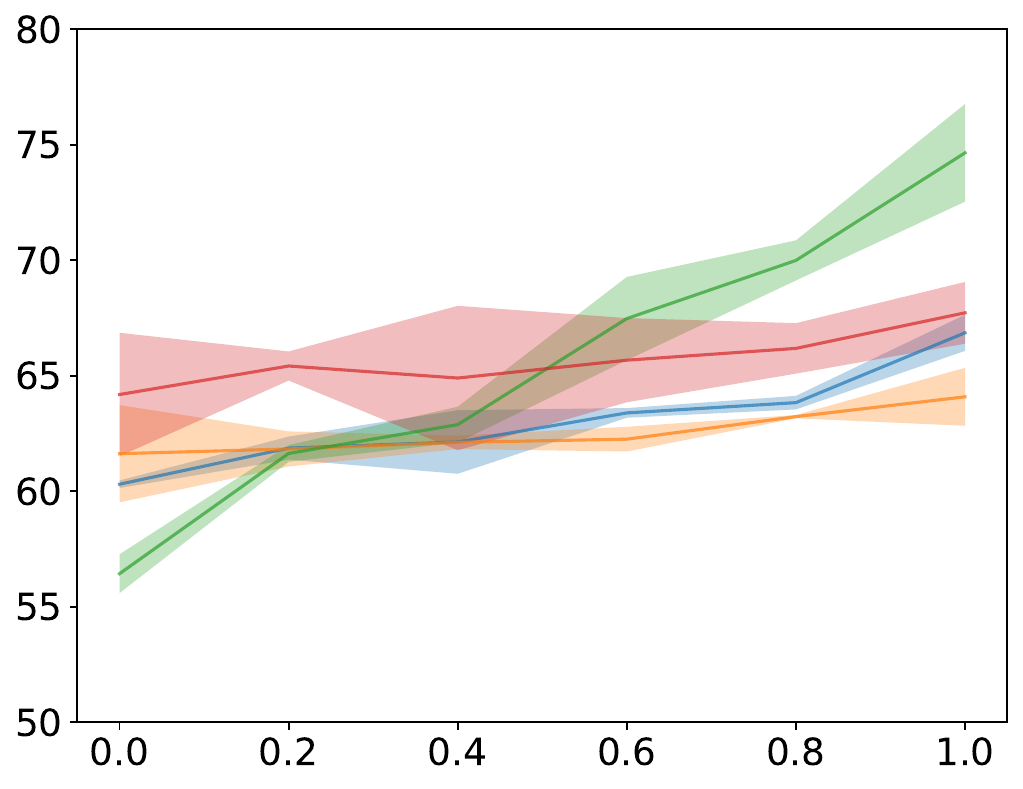} \\ [\tabvsize]

\end{tabular}

\caption{Affine Genetic Algorithm (AGA) parameter variation. We vary the AGA parameters $n_i$ (number of iterations), $p_c$ (crossover probability), $p_m$ (mutation probability), and $\epsilon$ (random noise intensity), and evaluate the results with the attack success rate (SR) score.}
\label{fig:results-genetic-param-var}
\end{figure}

\begin{figure*}[!hbt]
  \centering
  \includegraphics[width=1.032\columnwidth]{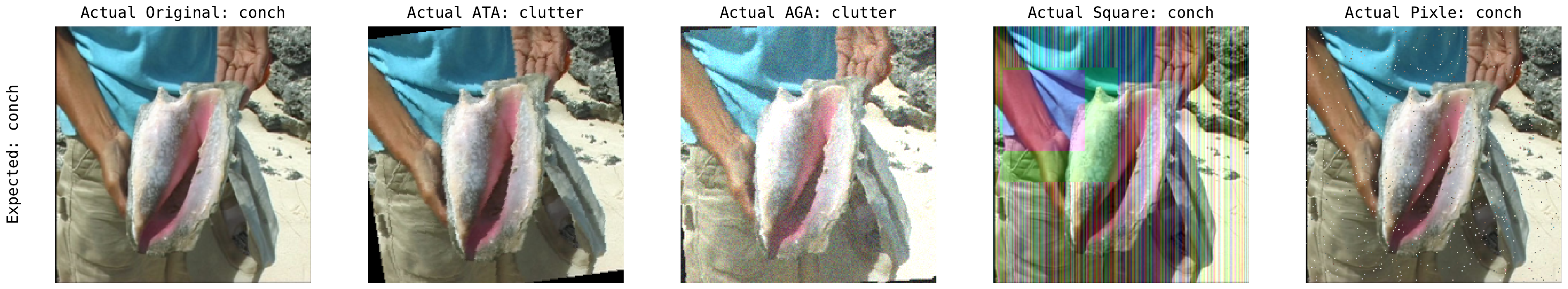}
  \includegraphics[width=1.032\columnwidth]{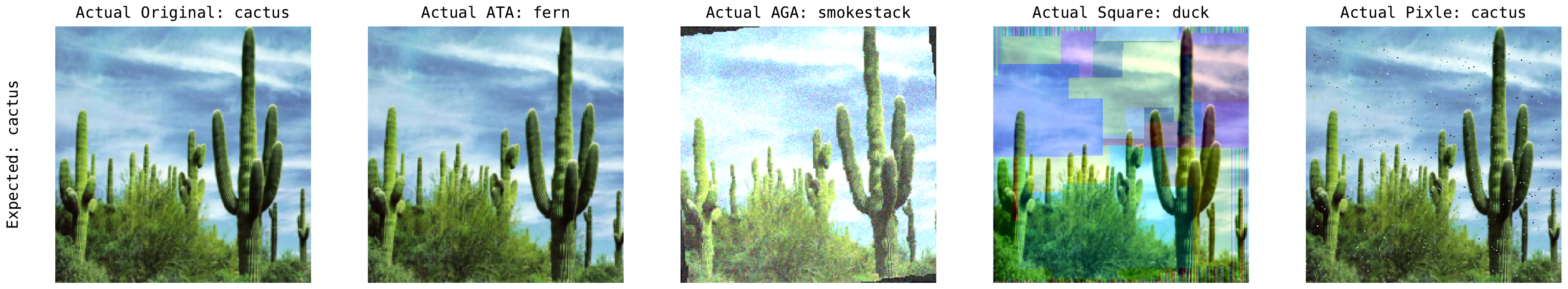}
  \includegraphics[width=1.032\columnwidth]{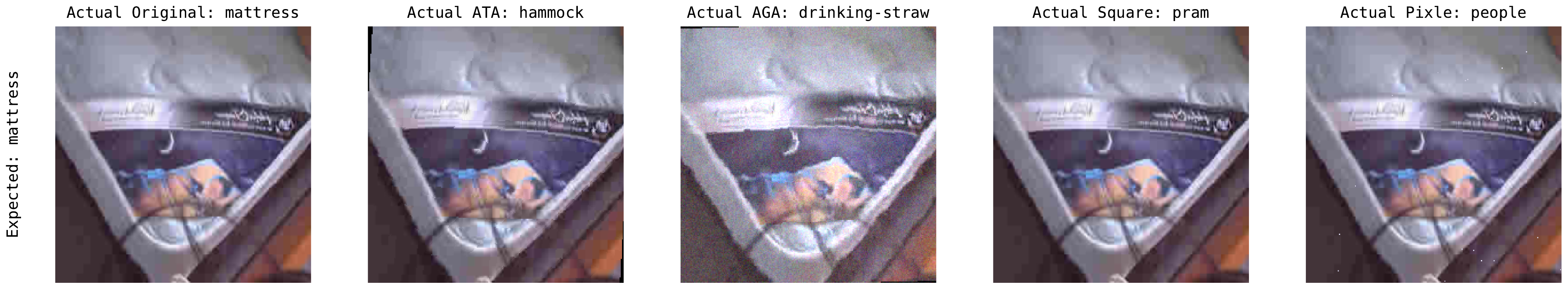}
  \includegraphics[width=1.032\columnwidth]{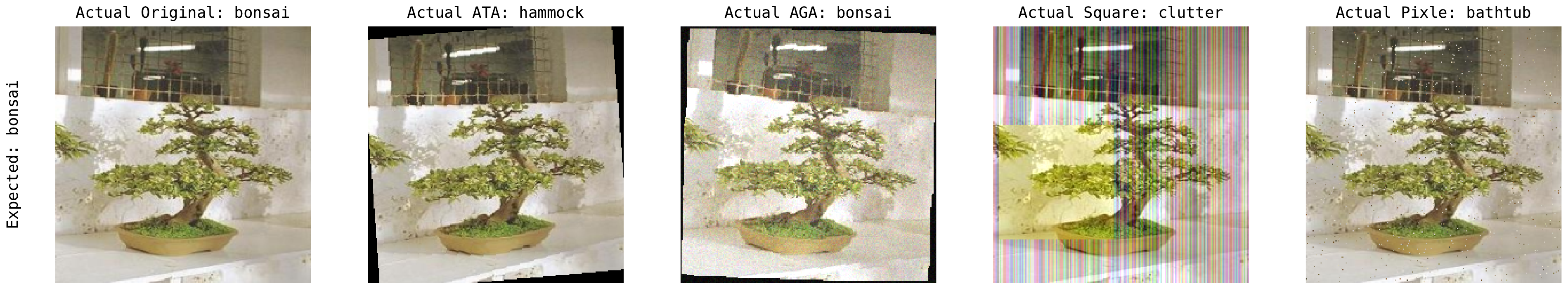}
  \includegraphics[width=1.032\columnwidth]{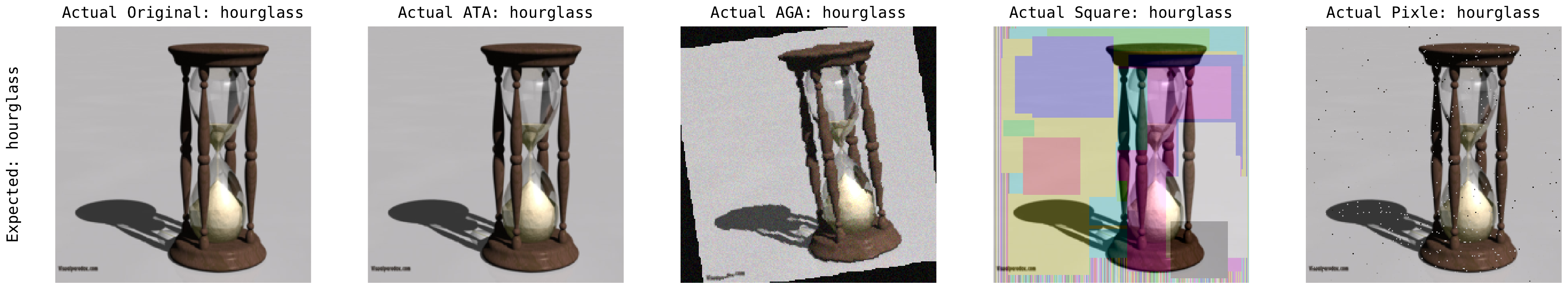}
  \includegraphics[width=1.032\columnwidth]{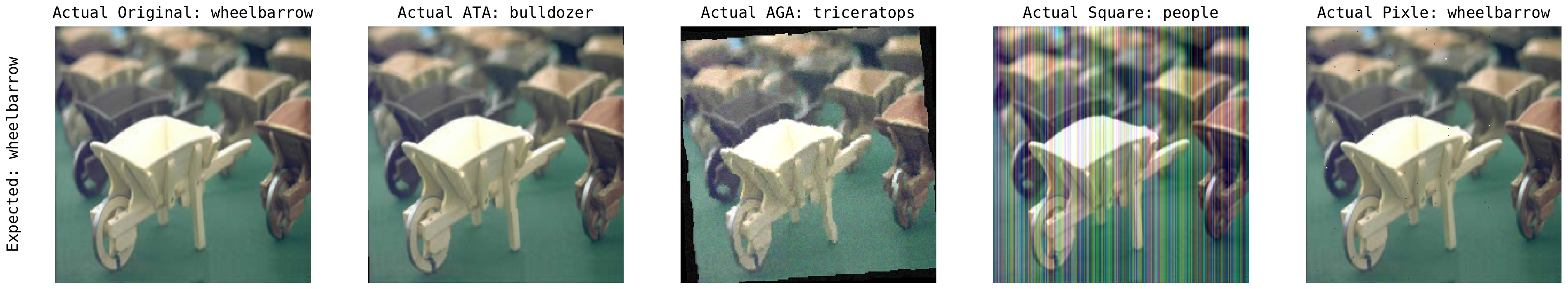}
  \includegraphics[width=1.032\columnwidth]{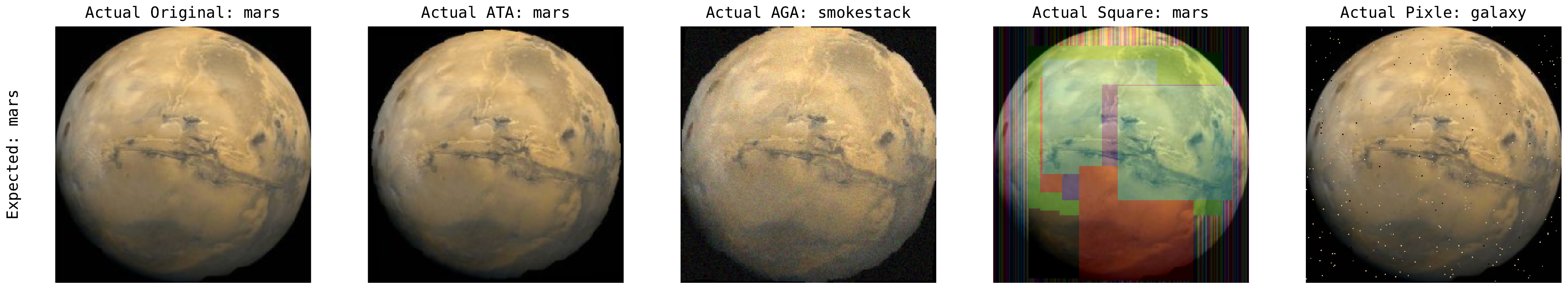}
  \includegraphics[width=1.032\columnwidth]{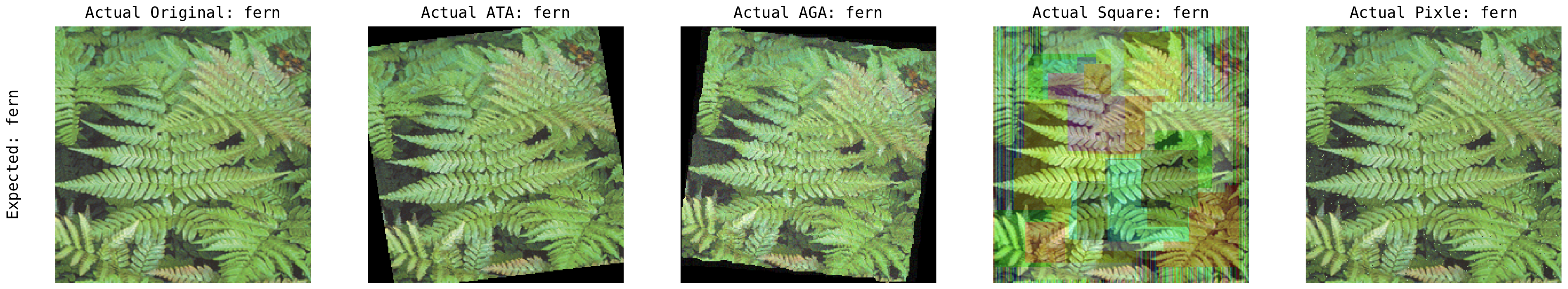}
  \includegraphics[width=1.032\columnwidth]{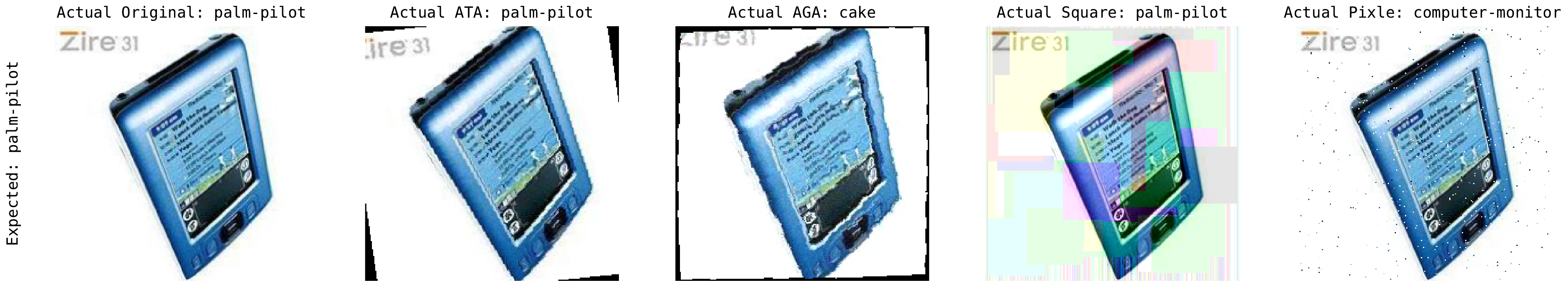}
  \includegraphics[width=1.032\columnwidth]{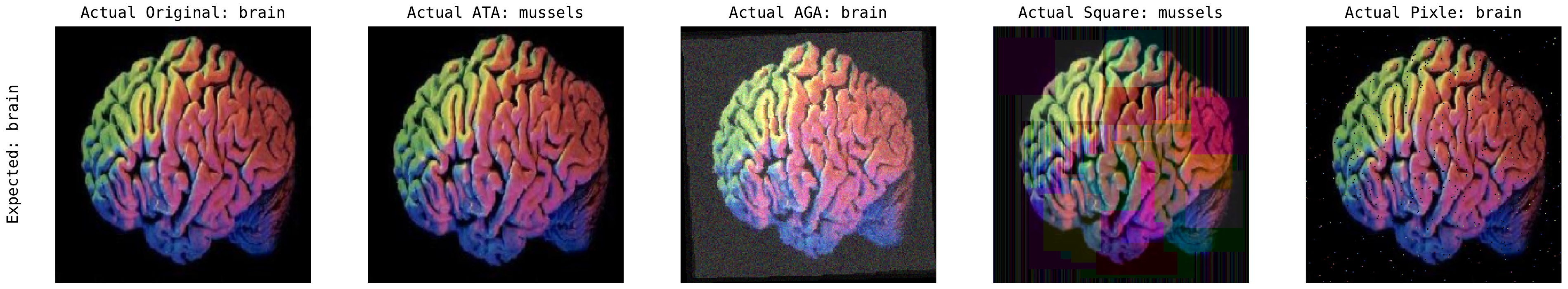}
  \includegraphics[width=1.032\columnwidth]{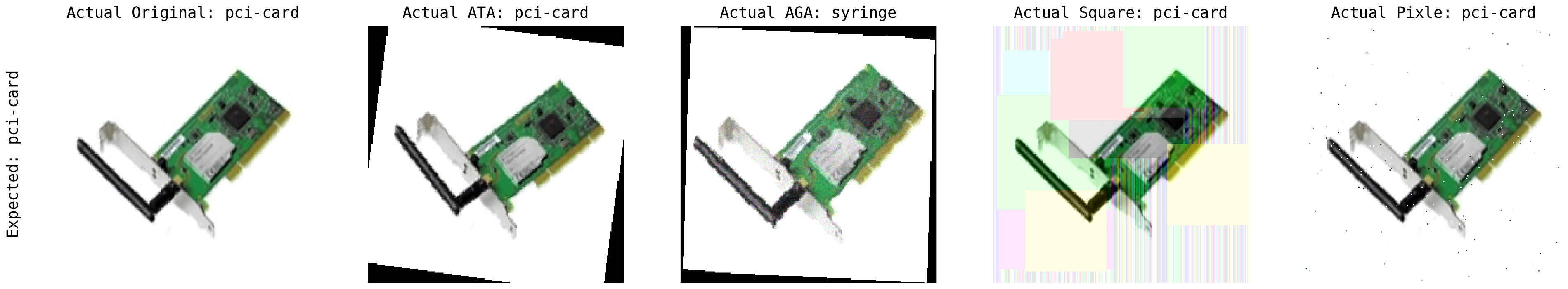}
  \includegraphics[width=1.032\columnwidth]{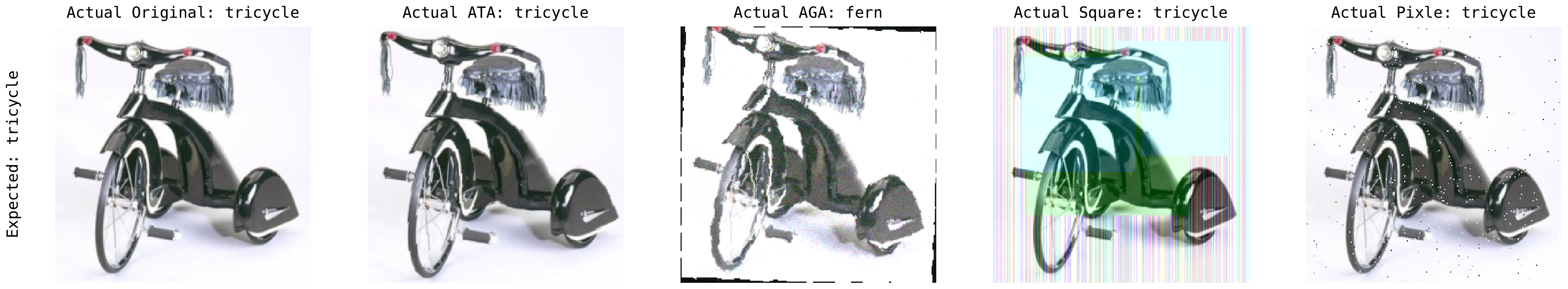}
  \caption{Comparison between ATA, AGA, Square Attack, and Pixle transformed examples and the original images. The Swin Transformer V2 Caltech-256 model evaluation results for each image are illustrated above the images as ``Actual Original'', ``Actual ATA'', ``Actual AGA'', ``Actual Square'', and ``Actual Pixle''. The ground truth is displayed to the left of each image series.}
   \label{fig:annex-ata-aga-square-pixle}
\end{figure*}

\begin{figure}[!htb]
\centering
\newcommand{\subfigsize}{0.155\textwidth}
\newcommand{\tabhsize}{\hspace{-2.5em}}
\newcommand{\tabvsize}{-2em}
\newcommand{\firstcolumnadjust}{-3.8}
\newcommand{\firstrowadjust}{1}
\tiny
\begin{tabular}{c >{\hspace{-2.5em}}  c >{\tabhsize} c >{\tabhsize} c >{\tabhsize} c}
    & Caltech-256 & Food-101 & Tiny-ImageNet-200  \\

    \multirow{-4}{*}{\rotatebox[origin=c]{90}{Attack SR (\%)}} & 
    \includegraphics[width=\subfigsize,valign=m]{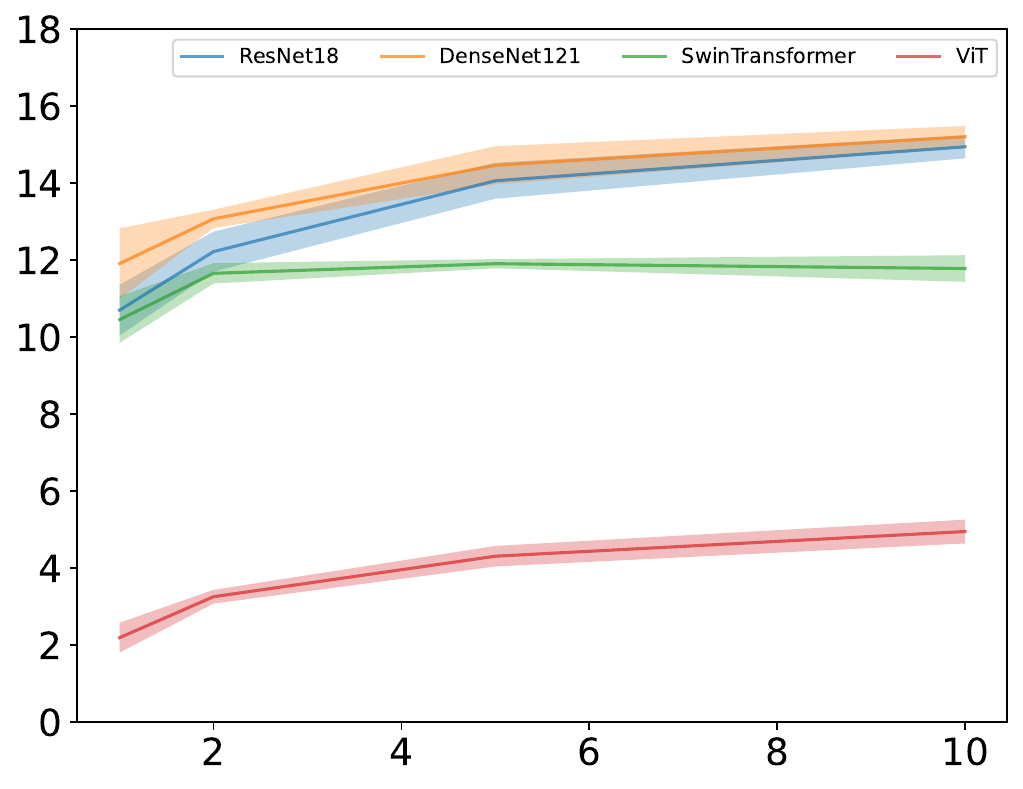} &
    \includegraphics[width=\subfigsize,valign=m]{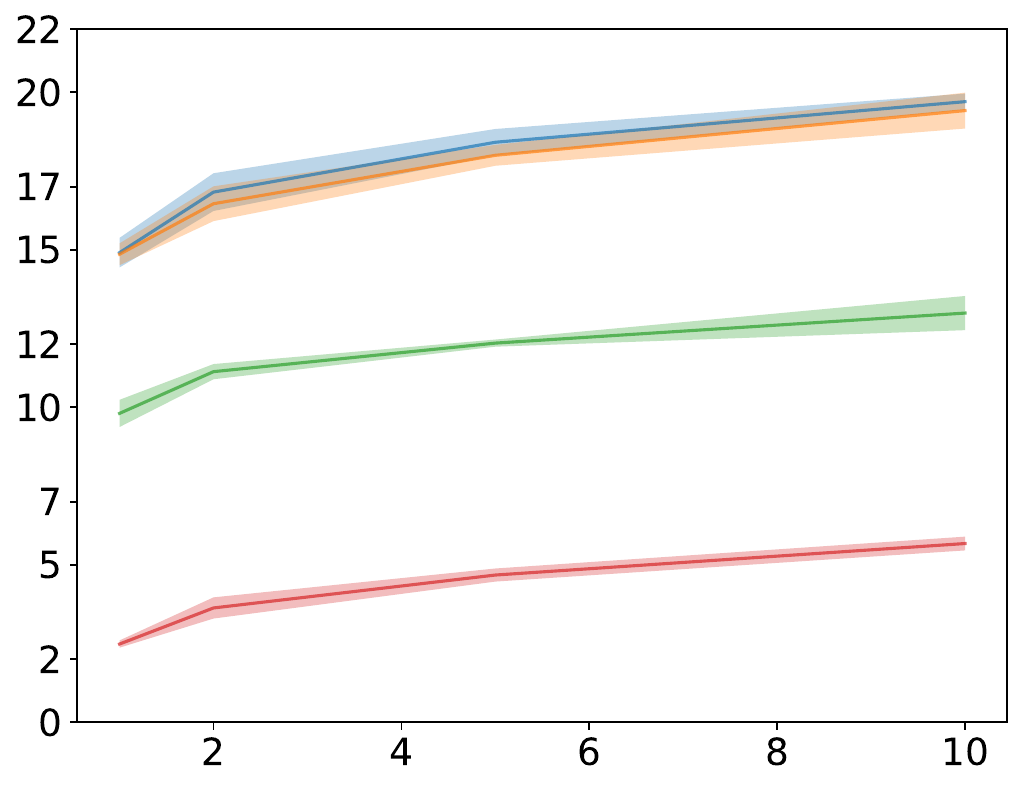} &
    \includegraphics[width=\subfigsize,valign=m]{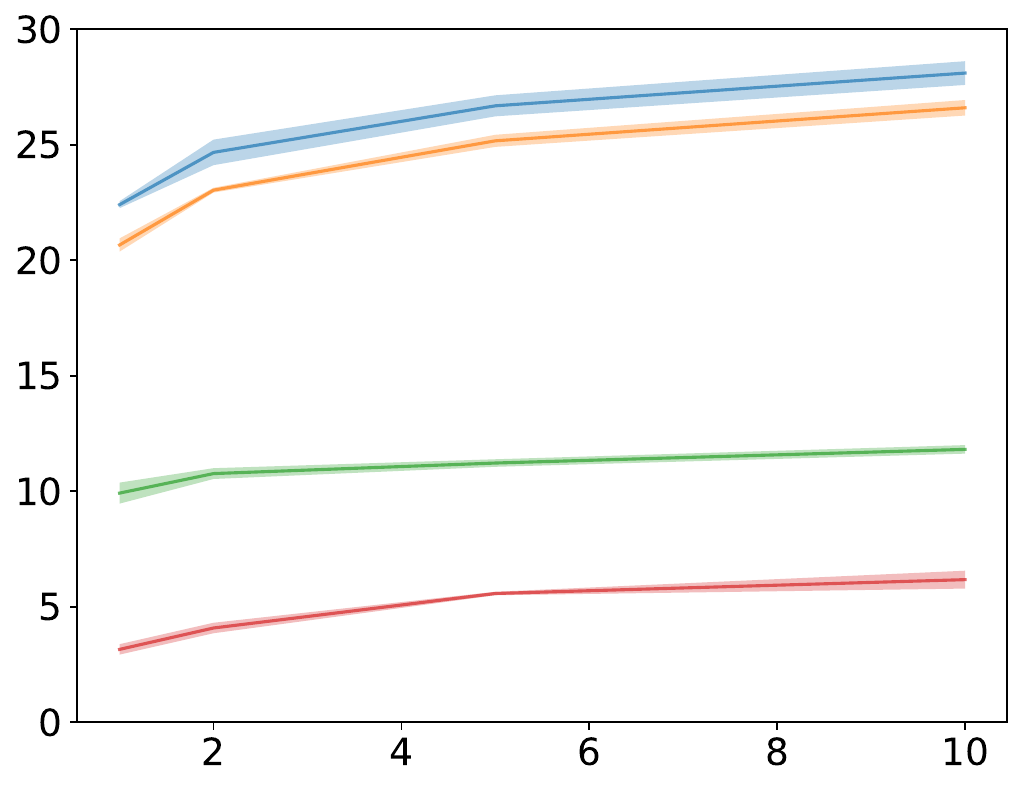} \\

\end{tabular}

\caption{Affine Transformation Algorithm (ATA) variation of the number of iterations ($n_i$) for each dataset. We evaluate the results with the attack success rate (SR) score.}
\label{fig:annex-ata-param-var}
\end{figure}

In \cref{fig:results-genetic-param-var} and \cref{fig:annex-ata-param-var}, the attack SR for ResNet-18 (ResNet18 label) is shown in blue, while DenseNet-121 (DenseNet121 label) is shown in orange, Swin Transformer V2 (SwinTransformer label) is shown in green, and Vision Transformer (ViT label) is in red.

\Cref{fig:results-genetic-param-var} illustrates the attack SR for each AGA algorithm parameter variation. Varying any parameter, we confirm the attack results in \cref{tab:results-untargeted-attack} by finding the Swin Transformer V2 as the most affected model for Caltech-256 and Food-101 datasets when the parameter values increase, and the ViT model as the most vulnerable to attacks for the Tiny-ImageNet-200 dataset. We find a constant growth in the SR for the $n_i$ parameter. However, the trend shows a potential peak for further increases in iterations. The average $p_c$ variation shows that the parameter has minimum sensitivity regardless of the population crossover probability. However, due to the significant randomness in the exchange of images between populations, we obtain up to a 7-8\% standard deviation. The mutation probability ($p_m$) exhibits an intuitive trend, with a peak at 100\% mutation probability, as all population examples are mutated through affine transformations and random noise. Finally, the intensity ($\epsilon$) applied to random noise shows a more pronounced impact of the attack as the value increases, with the Swin Transformer V2 being the most sensitive to its amplification.

Similarly with the results of the AGA algorithm parameter variation, \cref{fig:annex-ata-param-var} illustrates the number of iterations ($n_i$) of the ATA algorithm and the attack SR for each dataset. Unlike the AGA algorithm with up to 40-45\% added success, the SR score for each model and dataset is stable, with no significant increase on a higher $n_i$ value, featuring attack improvements of only 0.5-4\%. 
Moreover, the attack SR is visibly smaller for Swin Transformer V2 and ViT architectures. However, the former seems more vulnerable to attacks than ViT, with up to 6-8\% more attack SR. When comparing the results between datasets, the range bounds are proportionally consistent with the AGA results, with the highest attack success for Tiny-ImageNet-200 of around 27\%, followed by the Food-101 dataset, with almost 19\%, and Caltech-256, with around 15\% maximum attack SR.

\subsection{Qualitative Results}

We qualitatively assess our algorithms and compare the classification results on the Swin Transformer V2 model trained with the Caltech-256 data \cite{griffin2022caltech} without adversarial data augmentation. The comparison, depicted in \cref{fig:annex-ata-aga-square-pixle}, is performed between the original test input and the ATA, AGA, Square Attack \cite{andriushchenko2020squareattack}, and Pixle \cite{pomponi2022pixle} attacks.

Focusing on the qualitative differences between ATA and AGA, we notice the differences in the algorithm's iterative method. ATA applies rotation, scaling, translation, and shearing transformations once and chooses the best image candidate. In contrast, the AGA algorithm applies iterative transformations based on the steps of the genetic algorithm, updating the population with best-fit candidates; thus, edge smoothening and distortions are featured for some of the AGA output images. Additionally, AGA introduces noise to images, which accumulates iteratively from one generation to the next. Such an algorithmic contrast is better depicted in the ``mars'' image, where the planet's shape is deformed. In the ``fern'' and ``palm-pilot'' images, the image margins are slightly distorted and rounded. The cumulative noise of the AGA algorithm is best illustrated in the ``cactus'' images, where, compared to the ATA output, the AGA image is significantly brighter due to the additional noise added to it.

Comparing the ATA and AGA images with the Square Attack and Pixle, we notice that the Square Attack applies noise as square patches and noise stripes. Moreover, Pixle output is only an interchange of image pixels; thus, some added noise distorts the original images in the Square Attack and Pixle cases. A noticeable example for Pixle is the ``mars'' image, whereby, with a slight change of pixels, the image was classified as ``galaxy,'' fooling the model into perceiving that there are other background stars in the picture.

Quantitatively, considering the 12 core images depicted in \cref{fig:annex-ata-aga-square-pixle}, given that the model is trained on the dataset, the original image is correctly classified in 12/12 cases. Furthermore, adversarial success is obtained in 6/12 cases for ATA, in 8/12 cases for AGA, in 5/12 cases for Square Attack, and in 4/12 cases for Pixle. 

Finally, even the slightest image distortion or noise addition can fool the model. For ATA, the ``mattress'' image is perceived as ``hammock'', while the Square output of the same original image is classified as ``pram'', and as ``people'' for the Pixle resulting image. For other images, such as ``fern'', even the visible change in the picture results in the exact correct classification as for the original image.


\section{Conclusions}
This work provides a benchmark for two novel adversarial algorithms with noteworthy potential for several deep neural network improvement approaches. Based on data augmentation, we demonstrate that our algorithms outperform a similar method, as shown in the results depicted in \cref{tab:results-model-performance}. Additionally, we conduct a detailed analysis of both adversarial attacks and defenses. Despite the use of affine transformation and genetic algorithm randomness, the results in \cref{tab:results-untargeted-attack} demonstrate a significant defense against highly impactful attacks. Furthermore, the outcomes of targeted attacks, highlighted in \cref{tab:annex-targeted-attack}, illustrate the potential of adversarial attacks to influence models, with a focus on improving the classification of target classes. This enables further research into improving highly performant models with low accuracy in specific dataset classes through targeted data augmentation. Finally, we assess the consistency and robustness of the algorithms by varying their parameters and qualitatively compare the ATA, AGA, Pixle, and Square Attack black-box algorithms.

Among our main insights, we conclude that the computer vision transformer architectures (Swin Transformer V2 and ViT) achieve the best accuracy in regular training. However, they are significantly more vulnerable to undefended adversarial attacks based on the AGA algorithm; therefore, future robustness research should focus more on transformers. Adversarial training reverses this vulnerability, as computer vision transformers benefit disproportionately from adversarial augmentation and can become more robust than CNNs when trained on adversarial data. 

Furthermore, our AGA algorithm is the strongest attack method evaluated, producing more adversarially effective images. Its high performance is mainly driven by the number of iterations, mutation rate, and noise intensity, while the crossover rate has a limited impact on the attack efficacy. In contrast, the ATA algorithm is more stable and less sensitive to heavy tuning, making it more suitable for data augmentation or as part of a more complex adversarial algorithm.

Since we have used our algorithms for image classification, evaluating their potential for other tasks, such as natural language processing, scene text detection, and face recognition, is a subject for further work. Moreover, we adjusted the algorithm hyperparameters to the available hardware; thus, additional testing with increased parameter values is the subject of further research. In addition, for brevity, we limited targeted adversarial attacks to only ten classes for a dataset; a clearer picture of the benefits of targeted attacks will be drawn from the complete testing of multiple datasets. Finally, future research will also cover additional qualitative results, explainability investigations, exploration of multi-objective genetic algorithm techniques, affine transformation ablation studies, and comparisons with other existing white-, gray-, and black-box adversarial algorithms. Furthermore, the progression of the ATA and AGA algorithms toward white-box adversarial attacks warrants rigorous study and evaluation.



\begin{ack}
This work is supported by the project \textit{Romanian Hub for Artificial Intelligence - HRIA}, Smart Growth, Digitization and Financial Instruments Program, 2021-2027, MySMIS no. 334906, and by the National University of Science and Technology POLITEHNICA Bucharest through the PubArt program. 
\end{ack}


\bibliography{arxiv}

\end{document}